\documentclass{article}

\usepackage{arxiv}

\usepackage[utf8]{inputenc} 
\usepackage[T1]{fontenc}    
\usepackage{hyperref}       
\usepackage{url}            
\usepackage{booktabs}       
\usepackage{amsfonts}       
\usepackage{nicefrac}       
\usepackage{microtype}      
\usepackage{lipsum}		
\usepackage{graphicx}
\usepackage[numbers]{natbib}
\usepackage{doi}
\usepackage{longtable}

\usepackage{tikz}
\usetikzlibrary{positioning, fit, arrows.meta, shapes, chains}

\title{Deep Emotion Recognition in Textual Conversations: A Survey}

\author{ \href{https://orcid.org/0000-0002-9653-7181}{\includegraphics[scale=0.06]{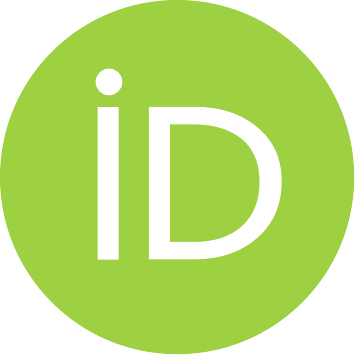}\hspace{1mm}Patrícia Pereira}\\
	INESC-ID and Instituto Superior Técnico,\\
	Universidade de Lisboa,\\
	Portugal \\
	\texttt{patriciaspereira@tecnico.ulisboa.pt} \\
	\And
	\href{https://orcid.org/0000-0003-0900-6938}{\includegraphics[scale=0.06]{orcid.pdf}\hspace{1mm}Helena Moniz}\\
	INESC-ID and Faculdade de Letras,\\
	Universidade de Lisboa,\\
	Portugal \\
	\texttt{helena.moniz@inesc-id.pt} \\
	\And
	\href{https://orcid.org/0000-0003-0005-8299}{\includegraphics[scale=0.06]{orcid.pdf}\hspace{1mm}Joao Paulo Carvalho}\\
	INESC-ID and Instituto Superior Técnico,\\
	Universidade de Lisboa,\\
	Portugal \\
	\texttt{joao.carvalho@inesc-id.pt} \\
}

\renewcommand{\undertitle}{}

\hypersetup{
pdftitle={Deep Emotion Recognition in Textual Conversations:
A Survey},
pdfkeywords={Emotion Recognition, Conversations, Deep Learning, Sentiment Analysis, Dialogue},
}

\begin{document}
\maketitle

\begin{abstract}
Emotion Recognition in Conversations (ERC) is a key step towards successful human-machine interaction. 
While the field has seen tremendous advancement in the last few years, new applications and implementation scenarios present novel challenges and opportunities. 
These range from leveraging the conversational context, speaker, and emotion dynamics modelling, to interpreting common sense expressions, informal language, and sarcasm, addressing challenges of real-time ERC, recognizing emotion causes, different taxonomies across datasets, multilingual ERC, and interpretability. This survey starts by introducing ERC, elaborating on the challenges and opportunities of this task. It proceeds with a description of the emotion taxonomies and a variety of ERC benchmark datasets employing such taxonomies. This is followed by descriptions comparing the most prominent works in ERC with explanations of the neural architectures employed. Then, it provides advisable ERC practices towards better frameworks, elaborating on methods to deal with subjectivity in annotations and modelling and methods to deal with the typically unbalanced ERC datasets. Finally, it presents systematic review tables comparing several works regarding the methods used and their performance. Benchmarking these works highlights resorting to pre-trained Transformer Language Models to extract utterance representations, using Gated and Graph Neural Networks to model the interactions between these utterances, and leveraging Generative Large Language Models to tackle ERC within a generative framework. This survey emphasizes the advantage of leveraging techniques to address unbalanced data, the exploration of mixed emotions, and the benefits of incorporating annotation subjectivity in the learning phase.
\end{abstract}


\section{Introduction}
\label{introduction}

This survey systematically reviews Deep Emotion Recognition in Conversations (ERC) works. It elaborates on several ERC challenges and opportunities, explaining how these are currently tackled by the reviewed works and discussing how these can be further addressed.

Emotion recognition capabilities are essential not only for successful interpersonal relationships but also for human-machine interaction. Understanding and knowing how to react to emotions significantly improves the interaction and its outcome. It is therefore a crucial component in the development of empathetic machines, which substantially enriches the experiences these can provide.

\begin{figure}[!ht]
	\begin{center}
		\includegraphics[width=0.5\linewidth]{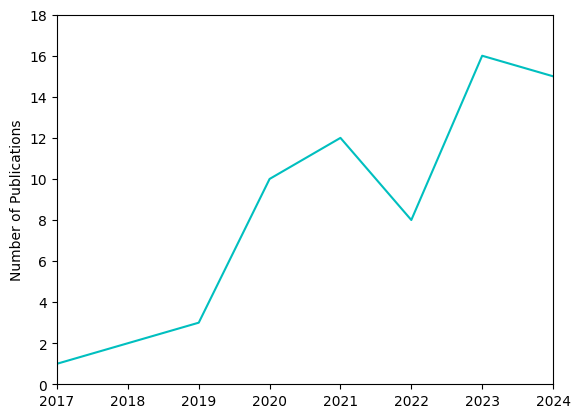}
		
	\end{center}
	\caption{Evolution of the number of ERC publications across the years}
	\label{plot}
\end{figure}

ERC modules are useful for a wide range of applications, from automatic opinion mining, to emotion-aware conversational agents and as assisting modules for therapeutic practices. ERC is therefore an actively growing research field, with high applicability and potential. Figure \ref{plot} depicts the evolution of the number of ERC publications across the years, using the most established benchmark datasets  \citep{busso2008iemocap, poria2018meld, zahiri2018emotion, li2017dailydialog}, supporting the previous observation. Note that it does not account for datasets addressing emotion causes, a recent research direction. 

It should be noted that emotions are only fully recognizable by the person having the emotion. Thus emotion recognition should be interpreted as \emph{``inferring an emotional state from observations of emotional expressions and behavior, and through reasoning about an emotion-generating situation''} \citep{picard2000affective}. 

Progress in ERC benefits from the advancements in Sentiment Analysis, particularly in dialogues, whose tasks include determining the polarity and emotion of text. Research in Sentiment Analysis has come a long way and there is a perception that the field has reached its maturity, but there are still aspects and sub-tasks that need to be thoroughly addressed to attain true sentiment understanding \citep{poria2020beneath}. Such a statement applies to ERC as well. Emotion and sentiment are used interchangeably throughout this survey, but one should bear in mind they have different meanings: emotion is a psychological state while sentiment is the mental attitude created through the existence of the emotion \citep{damasio2020sentir}. 

The conversational setting presents additional challenges and opportunities, besides the ones of Emotion Recognition, such as the presence of intricate dependencies between the emotional states of the subjects who participate in the conversation, and the use of context and common sense to express emotions, amongst others described in this survey.

\subsection{Task Definition}

The task of ERC consists of determining the emotion of each utterance in a textual conversation, for which speaker information is provided for each utterance.
Formally, given a sequence of $N$ utterances $[(u_1, p_1), (u_2, p_2), \dots, (u_N, p_N)]$, where each utterance $u_i$ is spoken by participant $p_i$, the goal is to predict the emotion label $e_i$ for each utterance $u_i$.

\subsection{Survey Contributions}

To the best of our knowledge, the existent surveys on Emotion Recognition in Conversations are by Poria et al \cite{poria2019emotion}, focusing on the challenges and advances in text ERC, pointing out conversational context modelling, speaker-specific modelling, the presence of emotion shift, multiparty conversations and the presence of sarcasm, and a more recent survey by Fu et al \cite{electronics12224714}, on multimodal ERC, that covers context modeling in conversations, speaker dependency, methods for fusing multimodal information and presents challenges such as real-world ERC and classification on the typical ERC imbalanced datasets.

Related relevant surveys covered sentiment analysis research challenges and directions \citep{poria2020beneath} and textual emotion recognition and its challenges \citep{deng2021survey}, the latter including a section on textual emotion recognition in dialogue, elaborating on utterance context modelling and dynamic emotion modelling. However, these are not specific to conversations, not specifying how the models and practices employed in general textual emotion recognition relate to conversations, nor covering several ERC pertaining challenges.

Our survey discusses additional challenges for this task, such as recognizing emotion causes, different taxonomies across datasets, multilingual ERC, and interpretability. It also extensively covers solutions to dealing with imbalanced ERC datasets and learning with subjectivity from annotators.

This is also the first survey presenting an extensive list of Deep Learning works in ERC, stating their methods, modalities, and performance across various datasets, while also describing key works on the topic. We provide comprehensive descriptions of Deep Learning methods from the Multi-Layer Perceptron, Recurrent Neural Networks, Long Short-Term Memory Networks, Gated Recurrent Units, Convolutional Neural Networks, Graph Neural Networks, and Attention Mechanisms to the Transformer, in the light of Emotion Recognition in Conversations. While these methods are transversal to all emotion recognition tasks, and therefore all emotion recognition surveys, some are more appropriate for ERC. Examples are Transformer-based architectures that are better at preserving long-term dependencies between sequences of utterances than recurrent neural networks.

We also set up an online repository\footnote{\url{http://github.com/patricia-pereira/deep-emotion-recognition-in-textual-conversations}} with the sources we present, organized into categories such as overviews, datasets, and works on ERC. 

\subsection{Survey Methodology}

We report the steps of our systematic review: identification, screening, eligibility, and inclusion.

In the identification step, to collect ERC works for Sections \ref{serc} and \ref{tables}, we search the accepted paper's sections of leading NLP conferences (ACL, EMNLP, NAACL, EACL, COLING, and Findings) and additional conferences (AAAI, INTERSPEECH, and ICASSP), using the title keywords ``emotion'' in combination with ``conversational'' or ``conversations''. Journal papers are also collected from IEEE Access and IEEE Transactions on Affective Computing.

We limit our search to papers published after 2017, given the significant advances in Deep Learning models since that time, particularly the introduction of the Transformer architecture \citep{vaswani2017attention} and pre-trained Transformer models such as BERT \citep{devlin2018bert} and RoBERTa \citep{liu2019roberta}.

In the screening step, the inclusion criteria are papers that address emotion recognition within conversations, report performance on one or more of the four main benchmark datasets listed in subsection \ref{data} (IEMOCAP, MELD, EmoryNLP, and DailyDialog), and report F1-scores, the adequate evaluation metric for emotion recognition in conversations, as elaborated in subsection \ref{unbalanced}. The exclusion criteria are papers not reporting results on the aforementioned benchmark datasets or not reporting F1-scores.

For Sections \ref{challenges}, \ref{tax-data}, and \ref{practices}, we conduct topic-specific searches using Google Scholar, employing keywords such as ``challenges in ERC'' or  ``data taxonomies in emotion recognition''. Search results are filtered by relevance and publication date to prioritize recent and pertinent research.

To complement our searches, we use tools such as Connected Papers to discover papers related to the ones already retrieved.

To assess paper eligibility we evaluate each paper's full text, focusing on the reported results and alignment with the survey scope, more specifically, the use of Deep Learning methods. 

\subsection{Outline}

This survey is organised as follows: Section \ref{introduction} introduces the topic of Emotion Recognition in Conversations and Section \ref{challenges} presents its challenges and opportunities. Section \ref{tax-data} describes the emotion taxonomies and a variety of ERC benchmark datasets employing such taxonomies. 
Section \ref{serc} provides descriptions of the most prominent works in ERC and explains the Deep Learning architectures employed. 
Section \ref{practices} describes advisable ERC practices towards better frameworks, elaborating on methods to deal with subjectivity in annotations and methods to deal with the typically unbalanced ERC datasets.
Section \ref{tables} consists of systematic review tables comparing several works regarding the methods used and their performance and provides insights into how the aforementioned works attempted to solve the ERC challenges presented in section \ref{challenges}. Section \ref{future} elaborates on future work directions. Finally, Section \ref{conclusion} presents the conclusions, limitations and key findings of the survey.

 	\section{ERC Challenges and Opportunities}
 \label{challenges}
 
 Research in ERC has come a long way and several challenges have been addressed, as described in this section. From context, to speaker and emotion shift modelling to interpreting common sense, informal language and sarcasm, Deep Learning architectures have demonstrated their potential to address these aspects, although there is room for improvement. Application scenarios such as real-time ERC pose additional challenges. In this section, we describe such challenges and point to the sections that are connected with each challenge, and in subsection \ref{revisited}
 we wrap up the challenges, elaborating on how they are currently being addressed. In Section \ref{future}, we suggest future work directions concerning partly unaddressed challenges.
 
 \subsection{Context, Speaker and Emotion Dynamics Modelling}
 
 While early works \citep{lee2009modeling, wollmer2010context} have tackled challenges from conversational context modelling, to speaker-specific modelling and emotion dynamics, resorting to, for example, gated networks, recent ERC approaches have leveraged the novel developments in Deep Learning, such as Transformers and Transformer-based Pre-trained Models to address these aspects. 
 
 Context modelling can be interpreted as leveraging information from previous or surrounding words or utterances. 
 Speaker-specific modelling leverages information about each speaker's utterances, capitalizing on the fact that an utterance belongs to a given speaker, being extremely important in multiparty conversations. 
 Emotion dynamics concerns the variations of the emotions of the participants in the conversation, being more informative but also harder to model than emotion shift. Emotions concern not only who express them but also the receiver. For example, more conscious and reflective people tend to be more sensitive regarding other's emotions.
 
 Concerning these challenges, gated and graph-based neural networks and transformer-based architectures are successful at modelling long range dependencies between words in utterances and relations between utterances. Key ERC works in Section \ref{serc} are described with regards to how these Deep Learning architectures were employed to address these challenges.
 
 \subsection{Common Sense, Informal Language and Sarcasm}
 
 Contextual embeddings from language models, such as BERT \citep{devlin2018bert} and RoBERTA \citep{liu2019roberta}, are pre-trained with large amounts of data, which makes them suitable to deal with informal language, sarcasm, and use of common sense to express emotions, although there is room for improvement. Most of the ERC works featured in this survey make use of these contextual embeddings. 
 
 Regarding common sense, commonsense knowledge bases such as COMET \citep{bosselut2019comet} were developed to aid in leveraging the meaning behind phrases employing such knowledge. Three key ERC works described in Section \ref{serc} make use of knowledge bases. These knowledge bases, however, are not specific to emotions, which could be an extension to consider.
 
 Concerning sarcasm, speaker-specific modelling can aid in identifying the sarcastic participants in the conversation, to more accurately classify their utterances \citep{poria2019emotion}. 
 
 \subsection{Real-Time ERC}
 
 Systems to be deployed in real-life conversational scenarios may require real-time emotion recognition \citep{eyben2010line}, which is challenging not only for modelling and recognition but also in annotation because the boundaries of each expression of emotion have to be simultaneously recognized as the emotion category \citep{picard2000affective}. Datasets such as the ones described in subsubsection \ref{other-data} feature annotations in continuous time, being more suitable for real-time ERC.
 
 \subsection{Recognizing Emotion Causes}
 \label{causes}
 
 Recognizing emotion causes, or what triggered the emotions in the conversation, is also an only recently addressed challenge \citep{poria2021recognizing, wang-etal-2023-generative-emotion, jeong-bak-2023-conversational, hu2024unimeec, hua2024causal, 10640186}. It involves resorting to datasets enriched with emotion causes, guiding classifiers to identify not only the emotion at stake but also what caused that emotion to arise. Addressing this challenge can greatly contribute towards a more complete emotional understanding. 
 
 \subsection{Different Taxonomies}
 
 With the use of different taxonomies and taxonomy types, as described in subsection \ref{tax} and reflected in the different emotion representations in the datasets in subsection \ref{data}, it is not straightforward to compare and establish connections between the different datasets, making it difficult, for example, to fine-tune the same ERC model with data from more than a few datasets in which adjustments can be made to their respective taxonomies. 
 
 \subsection{Multilingual ERC}
 
 Most of the benchmark datasets used to evaluate models and develop ERC applications are in English, as it can be observed in subsection \ref{data}. Translating the English conversations to the target application languages is not ideal since some emotion can be lost in translation. The collection and annotation of multilingual datasets for ERC would benefit research in this direction. Some dialogues involve code-switching, i.e., featuring several languages in the same dialogue or even utterance \citep{kumar-etal-2023-multilingual}.
 The cultural aspect of emotion expression/interpretation also poses a challenge, since emotion is not universally expressed or understood in the same way across different cultures.
 
 \subsection{Interpretability}
 \label{interpretability}
 A property lacking in most of the works resorting to deep learning architectures is interpretability. It is difficult to interpret the reasoning behind these models' decisions and identify what are the characteristic features of each class. Examining attention weights is a way of determining to which features the models were attending to while computing classifications, which can aid interpretability, although there are some issues raised in the community \citep{jain2019attention, wiegreffe2019attention}. Other methods can be proposed to provide more interpretability to the ERC pipeline, such as Fuzzy Fingerprinting Transformer Embedding Representations \citep{pereira2023fuzzy}.

 \section{ERC Taxonomies and Benchmark Datasets}
 \label{tax-data}
 
 In this section, a description of different taxonomies, i.e., ways of representing emotions, is provided, followed by the presentation of various datasets that resort to such taxonomies. The ``dialogues in the datasets reflect our daily communication way'' \citep{li2017dailydialog}, therefore encompassing several emotions across different taxonomies.
 
 \subsection{Emotion Taxonomies}
 \label{tax}

 One of the most important issues in the study of emotion is whether there are basic types of affect or whether affective states should be modelled as combinations of locations on shared dimensions \citep{zachar2012categorical}. These two dominant views corresponded to the categorical approach, defended by Panksepp \cite{panksepp2004affective}, and to the dimensional approach, proposed by Russell \cite{russell1980circumplex}, respectively.
 This is reflected in two main taxonomy approaches, that are described and compared in this section.
 The Plutchik wheel of emotions  \citep{plutchik1982psychoevolutionary} is also described.
 
 \subsubsection{Dimensional vs Categorical Approach}
\begin{figure}[!ht]
 	\begin{center}
 		\includegraphics[width=0.5\linewidth]{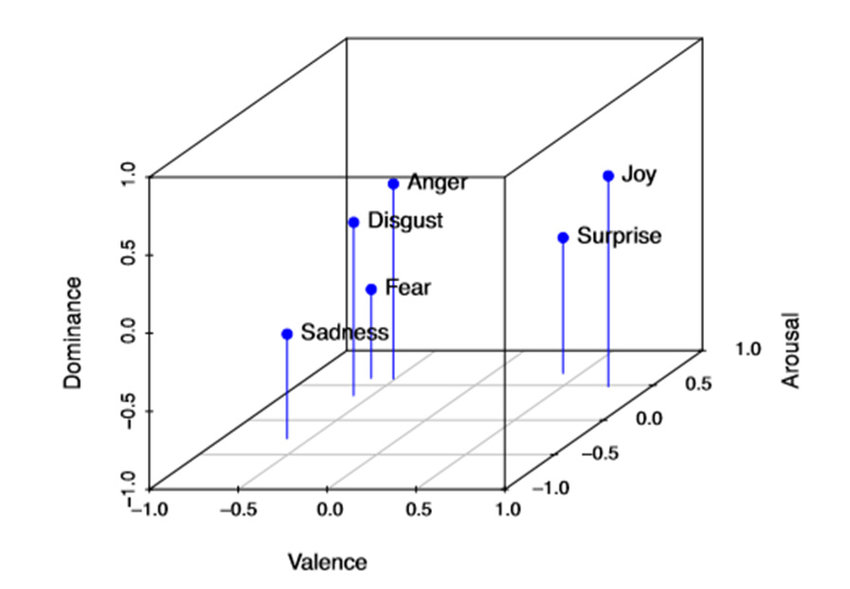}
 		
 	\end{center}
 	\caption{Distribution of common emotions over the VAD 3-dimensional space \citep{bualan2019emotion} }
 	\label{vad}
 \end{figure}

 The description of emotions can follow a dimensional approach, a continuous model that allows for an unlimited states' description. Following this approach, emotion is described by a range of values for valence, arousal, and dominance, the so-called VAD model \citep{mehrabian1980basic}. Valence is a measure of pleasure or displeasure of emotion: happiness has a positive valence, fear has a negative valence; arousal ranges from excitement to calmness: anger is high in arousal, sadness is low in arousal; dominance measures how much choice one has over an emotion: fear is low dominance, admiration is high dominance. The distribution of common emotions over the VAD 3-dimensional space can be observed in Figure \ref{vad}.

 A motivation for the use of the dimensional description could be the existence of mixed emotions, elaborated in subsubsection \ref{mixed}.
 
 The description can also follow a categorical approach, in which an emotion is assigned to one or more labels from a set of classes. Compared to the dimensional approach, this categorical discrete framework makes it easier to annotate emotions but is more limited. Several scholars proposed to identify a set of primary emotions. The basic emotions proposed by Ekman \cite{ekman1999basic} constitute a standard for this emotion description, categorised into joy, anger, sadness, surprise, fear, and disgust.

 \subsubsection{Plutchik Wheel of Emotions}
\begin{figure}[!ht]
 	\begin{center}
 		\includegraphics[width=0.5\linewidth]{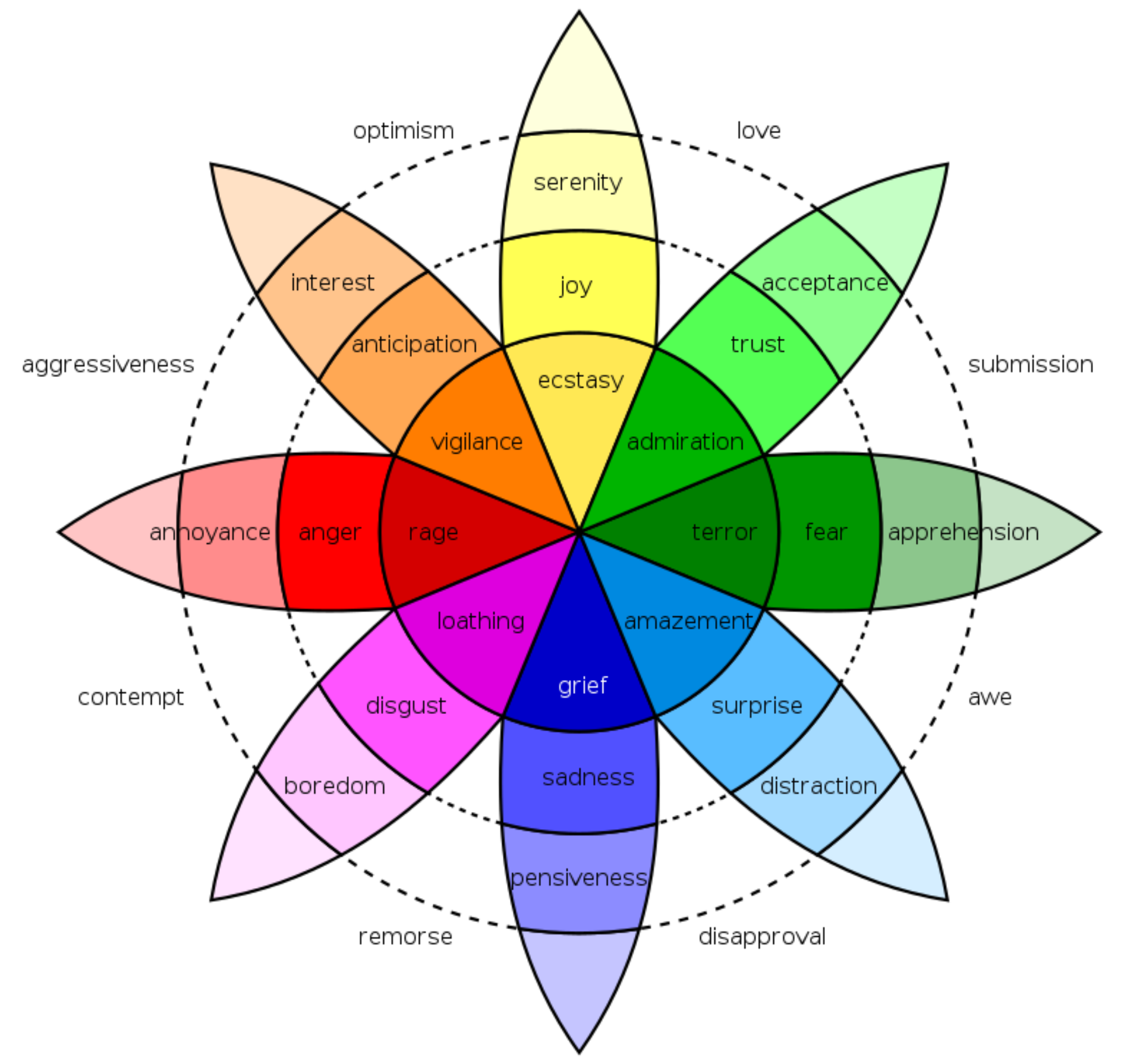}
 	\end{center}
 	\caption{The Plutchik wheel of emotions. Image retrieved from Wikipedia.}
 	\label{plut}
 \end{figure}

 The Plutchik wheel of emotions, depicted in Figure \ref{plut}, identifies eight core emotions, organised in opposite pairs of a wheel: sadness and joy, anger and fear, expectation and surprise, and trust and disgust. These have associated emotions according to their intensities and also combine to form new emotional states \citep{plutchik1982psychoevolutionary}.
 
 \subsubsection{Mixed Emotions}
 \label{mixed}
 A person can experience simultaneously several emotions. For example, one can experience joy and surprise at the same time and sometimes even opposite emotions simultaneously \citep{berrios2015eliciting}. While the VAD model allows for an unlimited description of emotions so that a mixture of emotions can be expressed in a single VAD description, the single-label categorical approach does not account for mixed emotions. This motivates the need to classify emotions with multiple labels when using the categorical approach. This need becomes more obvious with the decreasing number of labels in a dataset, but there are also multi-label datasets with a high number of emotion labels, such as the GoEmotions dataset \citep{demszky2020goemotions} with 28 emotions. 
 
 \subsubsection{ERC Taxonomy Considerations}
 
 The existent taxonomies are usually transversally applied along all Emotion Recognition tasks. However, for ERC, it can be observed that there are coincident predominant classes in several benchmark datasets, such as Neutral and Happiness/Joy, as can be seen in the class distribution of the datasets on subsection \ref{data}. Within different ERC domains such as chit-chat vs task-oriented, e.g. mental health \citep{DHEERAJ2021115265} and customer support \citep{herzig-etal-2016-classifying}, there are specific classes of emotions that will be more represented than others in each domain. With such a rich diversity of possible classes for this task, as it can be observed in the 28 classes of the GoEmotions dataset, the dimensional approach can encompass all classes but might make it harder to distinguish between differentiated emotional displays.

 \subsection{ERC Benchmark Datasets}
 \label{data}
 In this subsection, different categorical, single-label datasets are presented, that were used by the works benchmarked in this survey. These works are organised in Table \ref{ercmul}. Other relevant datasets are also presented. Overall, their class imbalance is significant suggesting the use of balancing techniques that will be elaborated on in subsection \ref{unbalanced}. All datasets are in English, which is a limitation concerning multilingual ERC, as pointed out in section \ref{challenges}.
 
 \subsubsection[IEMOCAP]{IEMOCAP \citep{busso2008iemocap}}
 The IEMOCAP dataset consists of 151 videos of two speakers’ dialogues, comprising 7.433 utterances with an average length of 11,6 words. Each utterrance is labelled with one of 8 emotions: happy, sad, neutral, angry, excited, frustrated fear, and disgust. Most of the works, however, benchmark their performance on the first 6 classes. The maximum class imbalance ratio is 1:3 (happy: frustrated).

 \vspace{2.5mm}
 \textbf{Sample:} \textit{``Is that, is that - is that just foam?  I can't even tell.  Although, if you can't tell, it's probably isn't them.  It'll probably be unmistakable, don't you think?''} \textbf{Label:} Excited

 \subsubsection[MELD]{MELD \citep{poria2018meld}}
 
 The Multimodal EmotionLines Dataset (MELD) contains 13.708 utterances from 1.433 dialogues from the TV series Friends. Each utterance encompasses audio, visual, and textual modalities, with an average length of 8,0 words. Emotion categories are Ekman's basic emotions and neutral. The maximum class imbalance ratio is 1:18 (fear: neutral).

 \vspace{2.5mm}
 \textbf{Sample:} \textit{``Are we okay now?''} \textbf{Label:} Fear

 \subsubsection[EmoryNLP]{EmoryNLP \citep{zahiri2018emotion}}
 
 EmoryNLP is a text dataset extracted from transcripts of the TV show Friends. It contains 12.606 utterances, labelled with one of the six primary emotions in Willcox's feeling wheel \cite{willcox1982feeling} and a neutral label. The maximum class imbalance ratio is 1:4 (sad: neutral).

 \vspace{2.5mm}
 
 \textbf{Sample:} \textit{``What a coincidence, I listen in my sleep.''} \textbf{Label:} Peaceful
 
 \subsubsection[DailyDialog]{DailyDialog \citep{li2017dailydialog}}
 
 DailyDialog is a large text dataset built from websites used to practise English dialogue in daily life, containing 13.118 multi-turn dialogues, comprising 102.879 utterances with an average length of 14,6 words. Each utterance is labelled with one of the six Ekman’s basic emotions or other. The maximum class imbalance ratio is 1:1156 (fear:other).

 \vspace{2.5mm}

 \textbf{Sample:} \textit{``I was scared stiff of giving my first performance''} \textbf{Label:} Disgust

 Table \ref{datasets} shows a comparison of the label distributions of the aforementioned datasets and Table \ref{datasets_perc} shows a comparison of the corresponding percentages.
 
 \begin{table}[!htpb]
 	\caption{Label distribution comparison of different ERC datasets.}
 	\centering

 		\begin{tabular}{|c|c|c|c|c|}
 			
 			\hline
 			\textbf{Label} & \textbf{IEMOCAP} & \textbf{MELD} & \textbf{Emory NLP} & \textbf{Daily Dialog} \\
 			\hline
 			Neutral         & 1708  & 6436  & 3776  & 85572   \\
 			Happiness/Joy   & 648   & 2308  & 2755  & 12885   \\
 			Surprise        & -     & 1636  & -     & 1823      \\
 			Sadness         & 1084  & 1002  & 844   & 1150    \\
 			Anger           & 1103  & 1607  & 1332  & 1022   \\
 			Disgust         & -     & 361   & -     & 353      \\
 			Fear            & -     & 358   & 1646  & 74         \\
 			Frustration     & 1849  & -     & -     & -         \\
 			Excitement      & 1041  & -     & -     & -         \\
 			Peace           & -     & -     & 1190  & -         \\
 			Powerful        & -     & -     & 1063  & -         \\
 			\hline
 			
 	\end{tabular}
 	\label{datasets}
 \end{table}
 
 \begin{table}[!htpb]
 	\caption{Label percentage distribution comparison of different ERC datasets.}
 	\centering

 		\begin{tabular}{|c|c|c|c|c|}
 			\hline
 			\textbf{Label} & \textbf{IEMOCAP} & \textbf{MELD} & \textbf{Emory NLP} & \textbf{Daily Dialog}  \\
 			\hline
 			Neutral         & 23.0  & 47.0  & 30.0  & 83.2 \\
 			Happiness/Joy   & 8.7   & 16.8  & 21.9  & 12.5  \\
 			Surprise        & -     & 11.9 & -     & 1.8     \\
 			Sadness         & 14.6  & 7.3  & 6.7   & 1.1   \\
 			Anger           & 14.8  & 11.7  & 10.6  & 1.0    \\
 			Disgust         & -     & 2.6   & -     & 0.3        \\
 			Fear            & -     & 2.6  & 13.1 & 0.1        \\
 			Frustration     & 24.9  & -     & -     & -          \\
 			Excitement      & 14.0  & -     & -     & -         \\
 			Peace           & -     & -     & 9.4  & -        \\
 			Powerful        & -     & -     & 8.4  & -         \\
 			\hline
 	\end{tabular}
 	\label{datasets_perc}
 \end{table}

 \subsubsection{Other Datasets}
 \label{other-data}
 
 The SEMAINE dataset \citep{mckeown2011semaine} consists in audiovisual recordings of interactions between a human and an operator undertaking the role of an avatar with four personalities. Its five core emotion dimensions are valence, activation, power, anticipation/expectation, and intensity. 
 
 The RECOLA dataset \citep{ringeval2013introducing} is a multimodal corpus of spontaneous human interactions collected from a collaborative task. It includes not only audio and video, but also electrocardiogram and electrodermal activity data. Emotional behaviours are annotated in terms of arousal and valence, in continuous time, and social behaviours are in terms of agreement, dominance, engagement, performance, and rapport, in discrete time.
 
 The SILICONE dataset \citep{chapuis-etal-2020-hierarchical} consists of preexisting datasets which have been considered challenging and interesting. It has annotations on Dialog Acts and Sentiment/Emotion. Concerning the latter, it comprises DailyDialog, MELD, IEMOCAP and SEMAINE.
 
\section{Deep Learning for Emotion Recognition in Conversations}
\label{serc}

In this section, key ERC works are described, focusing on the challenges and opportunities of the conversational scenario, presented in Section \ref{challenges}, and on the diverse Deep Learning architectures that can be employed for the task. Descriptions of key Deep Learning models are provided along with descriptions of works that resorted to those models.  
The works on Emotion Recognition in Conversations using Deep Learning described in this survey were chosen to be representative of the different techniques that are employed, covering gated Neural Networks with and without bidirectionality, Memory Networks, Graph Neural Networks, Attention Mechanisms, and Transformers. 
It is remarkable to observe how in four years research evolved from a simple Long Short-Term Memory Network (LSTM), taking into account the context of the conversation, to complex gated and graph neural network architectures and how the invention of transformers was reflected in not only word embeddings, such as BERT, but also in new classifier architectures. 
Comparisons between these works are performed based on the reported weighted-F1-score without the majority class on the IEMOCAP dataset, the choice of most authors including the prominent works of Poria et al \cite{poria2017context}, Hazarika et al \cite{hazarika2018conversational}, Majumder et al \cite{majumder2019dialoguernn}, Ghosal et al \cite{ghosal2019dialoguegcn},  Zhong et al \cite{zhong2019knowledge}, Shen et al \cite{shen2020dialogxl}, Ghosal et al \cite{ghosal2020cosmic}, Wang et al \cite{wang2020contextualized}, Li et al \cite{li2021past}, Shen et al \cite{shen2021directed} and Lei et al \cite{lei2023instructerc}, since the dataset is not too imbalanced.

\subsection{Convolutional Neural Networks for Feature Extraction}

Convolutional Neural Networks (CNNs) are a type of feedforward neural network inspired by the biology of the visual cortex that mimics the local filtering over the input space performed by the receptive fields. In the case of text, a linear layer is applied to a sequence of words, each represented by its embeddings, separately and in the same way for each sequence. The convolutional layer can have many filters, each extracting a different feature, and is usually followed by a pooling operation. 
In the case of Emotion Recognition in Conversations, convolutional layers act as feature extractors, useful since strong predictors of the label can appear in different places in the input. With the advent of Pre-Trained Language Models for utterance representation, elaborated on subsection \ref{embeddings}, convolutional neural networks have been less used.  

\subsection{The Multi-Layer Perceptron}

Emotion Recognition is a classification task. The most simple neural network architecture that can be employed for such a task is the Multi-Layer Perceptron.

The Multi-Layer Perceptron (MLP) is a non-linear function approximator that can be used for classification and regression. It is a class of feed-forward artificial neural networks that is composed of at least three layers: an input layer, a hidden layer, and an output layer. Its complexity and representation power increases with the number of layers. A MLP is mathematically described by the following equation:

\begin{equation}\label{eq21}
	\hat{y}=f_{L}(W_{L}(\ldots f_{2}(W_{2}(f_{1}(W_{1}x+b_{1})+b_{2})\ldots )+b_{L},
\end{equation}

In which L is the number of layers, $W_l \in R^{l-1} \times R^{l}$ and $b_l \in \mathbb{R}^{l}$ are the weight matrix and bias vector corresponding to layer $l$, and $f_{l}$ is a non-linear activation function corresponding to layer $l$. Except for the input layer, all layers’ units are followed by a non-linear activation function, such as the sigmoid, the hyperbolic tangent, or the rectifier linear unit. When performing classification, the last layer is followed by the sigmoid activation function for binary classification or the softmax activation function for multiclass classification. A multiclass MLP with one hidden layer can is depicted in Figure \ref{mlp}.

\tikzset{%
	every neuron/.style={
		circle,
		draw,
		minimum size=1cm
	},
	neuron missing/.style={
		draw=none, 
		scale=2.5,
		text height=0.333cm,
		execute at begin node=\color{black}$\vdots$
	},
}

\begin{figure}[!ht]
	\begin{center}
		\begin{tikzpicture}[x=1.5cm, y=1.5cm, >=stealth]
			
			\foreach \m/\l [count=\y] in {1,2,missing,3}
			\node [every neuron/.try, neuron \m/.try] (input-\m) at (0,2.5-\y) {};
			
			\foreach \m [count=\y] in {1,missing,2}
			\node [every neuron/.try, neuron \m/.try ] (hidden-\m) at (1.5,2-\y) {};
			
			\foreach \m [count=\y] in {1,missing, 2}
			\node [every neuron/.try, neuron \m/.try ] (output-\m) at (3,1.5-\y*0.75) {};
			
			\foreach \l [count=\i] in {1,2,N}
			\draw [<-] (input-\i) -- ++(-1,0)
			node [above, midway] {$x_\l$};
			
			\foreach \l [count=\i] in {1,n}
			node [above] at (hidden-\i.north) ;
			
			\foreach \l [count=\i] in {1,C}
			\draw [->] (output-\i) -- ++(1,0)
			node [above, midway] {$\hat{y_\l}$};
			
			\foreach \i in {1,...,3}
			\foreach \j in {1,...,2}
			\draw [->] (input-\i) -- (hidden-\j);
			
			\foreach \i in {1,...,2}
			\foreach \j in {1,...,2}
			\draw [->] (hidden-\i) -- (output-\j);
			
		\end{tikzpicture}
		
	\end{center}
	\caption{A Multy Layer Perceptron with one hidden layer, input layer with feature dimension $N$ and output layer with feature dimension $C$.
	}
	\label{mlp}
\end{figure}
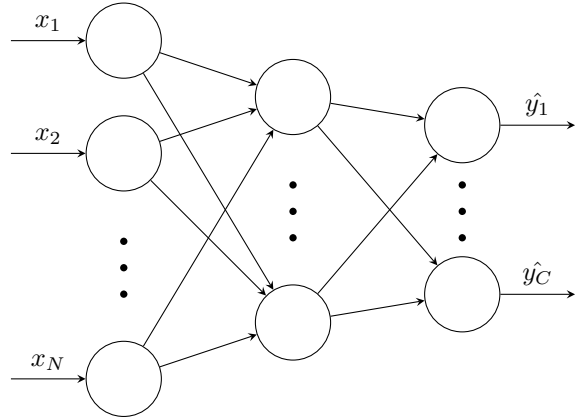

The MLP is commonly trained with the Backpropagation algorithm \citep{rumelhart1986learning}. The idea is to update the model parameters in the opposite direction of the gradient, such that a loss computed from the prediction error is minimised. The algorithm is composed of a forward pass and a backward pass. In the forward pass, the input is passed through all the layers in the network, yielding a prediction. In the backward pass, the gradients of the model’s parameters are computed with respect to the loss function, making use of the chain rule. These are computed one layer at a time, iterating backward from the last layer. The computation of the gradients along with weight update are done following an optimization method, commonly using Stochastic Gradient Descent (SGD) algorithms.

When used for Emotion Recognition, the input to the network is a set of extracted features representing each utterance obtained for example with the Bag of Words approach for text, which does not account for the order of the words, fitting this type of network but not being the ideal method for the task. That is the main reason why state-of-the-art works do not resort to this architecture.

\subsection{Elman Recurrent Neural Network}

Feed-forward neural networks, such as the above, assume data has no sequential dependencies which is not the case in Emotion Recognition in which an utterance is a sequence of words that can also comprise sequential data from other modalities.
Recurrent Neural Networks (RNNs) introduce memory into the network, being able to learn sequences. They process one input, such as a word, at a time, producing a hidden state which is a summary of the sequence of inputs up to the current timestep. The most simple one, the Elman RNN \citep{elman1991distributed}, updates the hidden state as in the following equation:

\begin{equation}\label{eq1}
	h_t=f(Wx_t+Uh_{t-1}+b),
\end{equation}
where $h_t \in \mathbb{R}^{h}$ is the hidden state at time $t$, $x_t \in \mathbb{R}^{N}$ is the input at time $t$, $W \in \mathbb{R}^{h\times N}$, $U \in \mathbb{R}^{h\times h}$ and $b \in \mathbb{R}^{h}$ are parameters to be learned and $f$ is a non-linear activation function such as the sigmoid or the hyperbolic tangent.
At each timestep the RNN can produce an output: 
\begin{equation}\label{eq3}
	\hat{y}_t=g(Vh_t+c),
\end{equation}
where $y_t \in \mathbb{R}^{y}$ is the output at time $t$, $ V\in\mathbb{R}^{y\times h} $ and $c \in \mathbb{R}^{y}$ are parameters to be learned and $g$ is a non linear activation function.

\begin{figure}[!ht]
	\begin{center}
		\begin{tikzpicture}[item/.style={circle,draw,thick,align=center}, itemc/.style={item,on chain,join}]
			\begin{scope}[start chain=going right,nodes=itemc, every join/.style={-latex,thick},local bounding box=chain]
				\path node (A0) {$W,U,b$} node[xshift=2em] (At)
				{$W,U,b$};
			\end{scope}
			\node[left=1em of chain,scale=2] (eq) {$=$};
			\node[left=2em of eq,item] (AL) {$W,U,b$};
			\path (AL.west) ++ (-1em,3em) coordinate (aux);
			\draw[thick,-latex,rounded corners] (AL.east) -| ++ (1em,3em) -- (aux) 
			|- (AL.west);
			\foreach \X in {0,t} 
			{\draw[thick,-latex] (A\X.north) -- ++ (0,2em)
				node[above] (h\X) {$h_\X$};
				\draw[thick,latex-] (A\X.south) -- ++ (0,-2em)
				node[below] (x\X) {$x_\X$};}
			\draw[white,line width=0.8ex] (AL.north) -- ++ (0,1.9em);
			\draw[thick,-latex] (AL.north) -- ++ (0,2em)
			node[above] {$h_t$};
			\draw[thick,latex-] (AL.south) -- ++ (0,-2em)
			node[below] {$x_t$};
			\path (x0) -- (xt) node[midway,scale=2,font=\bfseries] {\dots};
		\end{tikzpicture}
		
	\end{center}
	\caption{Two visual descriptions of the Recurrent Neural Network. The description on the left is unrolled, highlighting that parameters $W$, $U$ and $b$ are shared between timesteps.
	}
	\label{rnn}
\end{figure}
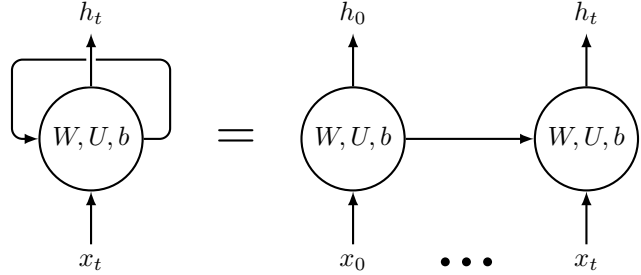

A visualization of the RNN update through time-steps is provided in Figure \ref{rnn}.

RNNs are trained with a modification of the Backpropagation algorithm, the Backpropagation Through Time (BPTT) algorithm \citep{werbos1990backpropagation}. All the parameters are shared between all time steps and it can be observed that an unrolled RNN is comparable to a feed-forward neural network with the same parameters shared over layers. 

In the case of Emotion Recognition, the input to the network is a sequential utterance, for example, the combination of text, audio, and visuals corresponding to one word being fed per timestep, and the output is the predicted emotion.

In some cases, particularly when dealing with long sequences such as when considering several utterances to model context in Emotion Recognition in Conversations, the backpropagation algorithm yields vanishingly small gradients, preventing the parameters from changing their values and eventually stopping training.

\subsection{Long Short-Term Memory Networks}

\begin{figure}[!ht]
	\begin{center}
		\resizebox{9cm}{8cm}{
			\begin{tikzpicture}[
				font=\sf \scriptsize,
				>=LaTeX,
				cell/.style={
					rectangle, 
					rounded corners=5mm, 
					draw,
					thick,
				},
				operator/.style={
					circle,
					draw,
					inner sep=-0.2pt,
					minimum height =.3cm,
				},
				function/.style={
					ellipse,
					draw,
					inner sep=1pt
				},
				ct/.style={
					circle,
					draw,
					line width = .75pt,
					minimum width=1cm,
					inner sep=1pt,
				},
				gt/.style={
					rectangle,
					draw,
					minimum width=4mm,
					minimum height=3mm,
					inner sep=1pt
				},
				mylabel/.style={
					font=\scriptsize\sffamily
				},
				ArrowC1/.style={
					rounded corners=.25cm,
					thick,
				},
				ArrowC2/.style={
					rounded corners=.5cm,
					thick,
				},
				]
				
				\node [cell, minimum height =4cm, minimum width=5cm] at (0,0){} ;
				
				\node [gt] (ibox1) at (-2,-0.75) {$\sigma$};
				\node [gt] (ibox2) at (-1.5,-0.75) {$\sigma$};
				\node [gt, minimum width=1cm] (ibox3) at (-0.5,-0.75) {Tanh};
				\node [gt] (ibox4) at (0.5,-0.75) {$\sigma$};
				
				\node [operator] (mux1) at (-2,1.5) {$\times$};
				\node [operator] (add1) at (-0.5,1.5) {+};
				\node [operator] (mux2) at (-0.5,0) {$\times$};
				\node [operator] (mux3) at (1.5,0) {$\times$};
				\node [function] (func1) at (1.5,0.75) {Tanh};
				
				\node[] (c) at (-3.5,1.5) {$c_{t-1}$};
				\node[] (h) at (-3.5,-1.5) {$h_{t-1}$};
				\node[] (x) at (-2.5,-3) {$x_t$};
				
				\node[] (c2) at (3.5,1.5) {$c_t$};
				\node[] (h2) at (3.5,-1.5) {$h_t$};
				\node[] (x2) at (2.5,3) {$h_t$};
				
				\draw [ArrowC1] (c) -- (mux1) -- (add1);
				\draw [->, ArrowC1] (add1) -- (c2);
				
				\draw [ArrowC2] (h) -| (ibox4);
				\draw [ArrowC1] (h -| ibox1)++(-0.5,0) -| (ibox1); 
				\draw [ArrowC1] (h -| ibox2)++(-0.5,0) -| (ibox2);
				\draw [ArrowC1] (h -| ibox3)++(-0.5,0) -| (ibox3);
				\draw [ArrowC1] (x) -- (x |- h)-| (ibox3);
				
				\draw [->, ArrowC2] (ibox1) -- (mux1);
				\draw [->, ArrowC2] (ibox2) |- (mux2);
				\draw [->, ArrowC2] (ibox3) -- (mux2);
				\draw [->, ArrowC2] (ibox4) |- (mux3);
				\draw [->, ArrowC2] (mux2) -- (add1);
				\draw [->, ArrowC1] (add1 -| func1)++(-0.5,0) -| (func1);
				\draw [->, ArrowC2] (func1) -- (mux3);
				
				\draw [->, ArrowC2] (mux3) |- (h2);
				\draw (c2 -| x2) ++(0,-0.1) coordinate (i1);
				\draw [-, ArrowC2] (h2 -| x2)++(-0.5,0) -| (i1);
				\draw [->, ArrowC2] (i1)++(0,0.2) -- (x2);
				
			\end{tikzpicture}
			
		}
		
	\end{center}
	\caption{Long-Short Term Memory Network cell}
	\label{lstm}
\end{figure}
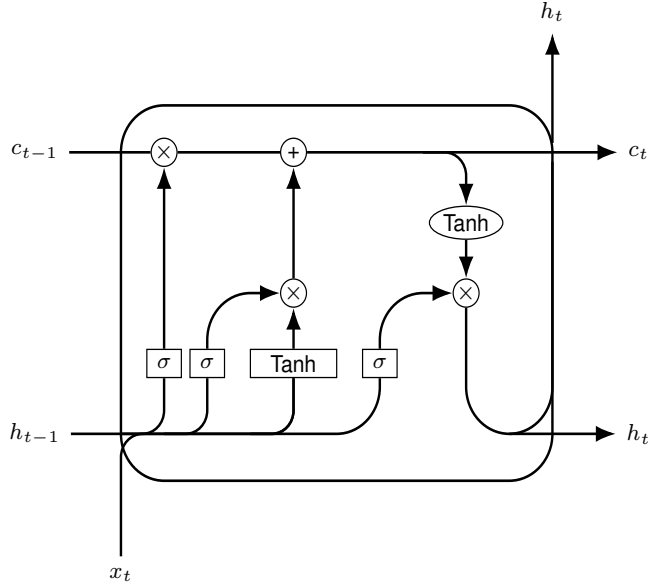

To overcome this problem, Long Short-Term Memory Networks (LSTMs) \citep{hochreiter1997long} were proposed. LSTMs are composed of a memory cell and three gates. The memory cell stores the information about the input sequence across timesteps while the gates control the flow of information across the memory cell: the input gate controls the proportion of the current input to include in the memory cell; the forget gate the proportion of the previous memory cell to forget; the information gate the information to output from the current memory cell. This can be described by the following equations:
\begin{equation}\label{eq2}
	i_t=sigm(W_ix_t+U_ih_{t-1}+b_i)
\end{equation}
\begin{equation}\label{eq4}
	f_t=sigm(W_fx_t+U_fh_{t-1}+b_f)
\end{equation}
\begin{equation}\label{eq5}
	o_t=sigm(W_ox_t+U_oh_{t-1}+b_o)
\end{equation}
\begin{equation}\label{eq6}
	c_t=f_t \odot c_{t-1}+i_t\odot tanh(W_cx_t+U_ch_{t-1}+b_c)
\end{equation}
\begin{equation}\label{eq7}
	h_t=o_t \odot tanh(c_t),
\end{equation}
where $i_t$ is the input gate, $f_t$ the forget gate, $o_t$ the output gate, $c_t$ the memory cell and $sigm$ and $tahn$ the element-wise sigmoid and hyperbolic tangent activation functions respectively. The several W, U and b are parameters to be learned. A visual depiction of an LSTM cell is provided in Figure \ref{lstm}.

The vanishing gradient problem is overcome by the gates, specially the forget gate.

As in the Elman RNN, in the case of Emotion Recognition the input to the network can be a sequential utterance and the output is the predicted emotion. LSTMs are usually leveraged to model dependencies between several sequential utterances. In this case, the input to the network is a sequence of utterances.

The LSTM described in the equations above does not capture information from future time steps, which is useful for example in text applications when the context of a word is made up of both previous and subsequent words in the phrase. Therefore, Bidirectional Long Short-Term Memory Networks (Bi-LSTM) \citep{graves2005bidirectional} were proposed. Bi-LSTMs consist of two LSTMs: one executing a forward pass and the other a backward pass. A sequence of hidden states is produced for each direction that are usually joined into a unique sequence of hidden states through concatenation.

\subsubsection[Context-Dependent Emotion Recognition]{Context-Dependent Emotion Recognition \cite{poria2017context}}

There is a strong correlation and influence in emotion distribution between sequential utterances in a conversation. Leveraging those contextual dependencies is key to the performance of a classifier. An utterance is a unit of speech which is bound by breathes or pauses \citep{olson1977utterance}. In textual data, it is usually delimited by punctuation such as full stops, but can also be delimited by question or exclamation marks.
Along these lines, the work of \cite{poria2017context} considers interdependences among utterances. The approach consists of an architecture made of LSTMs to extract contextual features from the utterances. The model enables consecutive utterances to share information while preserving their order. 
The authors propose the contextual LSTM, consisting of unidirectional LSTM cells, the hidden LSTM, in which the dense layer after the LSTM cell is omitted, and the bi-directional contextual LSTM that considers information from before and after the targeted utterance.
It achieved an F1-score of 54.95\% on the IEMOCAP dataset.

\subsection{Gated Reccurent Units}

Another RNN, which structure is somehow similar but with fewer parameters than the LSTM, is the Gated Recurrent Unit (GRU)  \citep{cho2014learning} that can be described by the following equations:

\begin{equation}\label{eq13}
	r_t=sigm(W_rx_t+U_rh_{t-1}+b_r)
\end{equation}
\begin{equation}\label{eq14}
	z_t=sigm(W_zx_t+U_zh_{t-1}+b_z)
\end{equation}
\begin{equation}\label{eq15}
	\hat{h}_t= tanh(W_hx_t+U_hh_{t-1}+b_h)
\end{equation}
\begin{equation}\label{eq16}
	h_t=z_t \odot h_{t-1}+(1-z_t)\odot\hat{h}_t,
\end{equation}

in which the $r_t$ is the reset gate that leads the candidate hidden state $\hat{h}_t$, to ignore or not the previous hidden state, and $z_t$ is the update gate that controls the proportion of the information from the previous hidden state to carry over to the current hidden state. 

GRUs can work better than LSTMs when there is less available data since they have fewer parameters, and are less prone to overfitting.

\subsubsection[DialogueRNN]{DialogueRNN \citep{majumder2019dialoguernn}}

DialogueRNN combines GRUs in a sophisticated way, in which each GRU plays a specific role in modelling the conversation. 

DialogueRNN predicts the emotion of an utterance based on the speaker, the context of preceding utterances, and the emotions from preceding utterances. The model has three components all modeled by GRUs. The party-state models the parties’ emotion dynamics throughout the conversation. The global state has the encoding of preceding utterances and the party state, and models the context of the utterance. Finally, the emotion representation is based on the party state and the global state and is used to perform the final classification. 

Several variants of the method are proposed. DialogueRNN$_l$ considers an additional listener state while a speaker utters, BiDialogueRNN uses a Bi-directional RNN,	DialogueRNN+Att uses attention over all surrounding emotion representations and BiDialogueRNN+Att combines the latter approaches.

It yielded an F1-score of 62.75\% on the IEMOCAP dataset. The authors argue DialogueRNN variants outperform the contextual LSTM due to better context representation.

\subsection{Attention Mechanisms}

The aforementioned work mentions the use of attention over surrounding emotion representations. To motivate Attention Mechanisms (AM) \citep{bahdanau2014neural} we now introduce Sequence to Sequence (Seq2Seq) models. Seq2Seq models are composed of an RNN encoder that takes the input sequence and outputs a context vector, the last encoder hidden state, that is fed into an RNN decoder that generates an output sequence from this vector. When the input sequence is very long the context vector does not have enough capacity to represent the information from the entire sequence. This is where AMs are useful. Taking inspiration from the human visual system that attends to different parts of the space, while building its representation of the scene, the AM allows the decoder to attend to the relevant encoder hidden states. At each time step, a context vector is obtained by a weighted sum of these hidden states, where the weights of the sum are proportional to the similarity between the current decoded hidden state and the encoded hidden states, as described in the following equations:

\begin{equation}\label{eq11}
	a_{ti}=\frac{exp(score(h_t^d,h_i^e))}{\sum_{j=1}^{T}exp(score(h_t^d,h_j^e))}
\end{equation}
\begin{equation}\label{eq12}
	c_t=\sum_{j=1}^{T}a_{tj}h_j^e,
\end{equation}

where $h_t^d$ is the decoder hidden state and $h_i^e$ and $h_j^e$ are the encoder hidden states. The most common similarity score function is the dot product.
By introducing AM in an RNN the performance of a classifier usually increases.

\subsection{Memory Networks}

\subsubsection[Conversational Memory Network]{Conversational Memory Network \citep{hazarika2018conversational}}

The recurrent networks described before are limited in terms of long-range summarization since they rely on a sequential processing approach with implicit memory.
The Conversational Memory Network (CMN) accounts for this factor through the use of memory networks, which can better capture long-term dependencies and have attention models that can summarize specific details using explicit memory structures.

The work is based on the notion that emotional dynamics involve self and inter-speaker emotional influence. Separate histories for each speaker, consisting of the speaker's previous utterances, are modeled into memory cells of the Conversational Memory Network using GRUs. Memory networks provide a memory component that can be read from and written to and also perform inference. The CMN then employs an attention mechanism over historical utterances from each speaker to filter out relevant content for the current utterance. This mechanism is repeated for multiple hops of the network to subsequently classify the utterance.
CMN achieves an F1-score of 56.13\% on the IEMOCAP dataset.

\subsubsection[Interactive Conversational Memory Network]{Interactive Conversational Memory Network 
	\citep{hazarika2018icon}}

The Interactive Conversational Memory Network (ICON) is an improvement upon CMN that has an additional dynamic global influence module to model inter-personal emotional influence, a module that maintains a global representation of the conversation, a global state that is updated using a GRU operation on the previous state and current speaker’s history. 
ICON yields an F1-score of 58.54\% on the IEMOCAP dataset.

\subsection{Graph Neural Networks}

An alternative to gated neural networks that is also useful to capture long term dependencies is using graph neural networks.
Graph neural networks are based on graphs. The latter model a set of objects, the nodes, and their relationships, the edges.
While standard neural networks such as CNNs and RNNs cannot handle the graph input properly since they need a specific order for the nodes, the output of GNNs is invariant for the input order of nodes. Furthermore, while in standard neural networks, dependencies between nodes are a node feature, in GNNs there are edges to represent these dependencies. GNNs can propagate information by the graph structure instead of using it as part of features \citep{zhou2020graph}.

Two common types of GNNs are recurrent GNNs (RecGNNs) and convolutional GNNs (ConvGNNs).  RecGNNs learn node representations with recurrent neural architectures, assuming that a node in a graph constantly exchanges information with its neighbors until a stable equilibrium is reached. ConvGNNs generalize the convolution operation from grid structure to graph structure, generating a node representation by aggregating its features with its neighbours' features. 
For classification tasks, such as Emotion Recognition, an end-to-end framework can be constructed for example by stacking graph convolutional layers followed by a softmax layer \citep{wu2020comprehensive}.

\subsubsection[DialogueGCN]{DialogueGCN \citep{ghosal2019dialoguegcn}}

Dialogue Graph Convolutional Network (GCN) takes into account intra and inter-speaker dependency and background information, having a sequential and a speaker-level encoder. The sequential context encoder encodes the utterances using a bidirectional GRU and is speaker-agnostic. The speaker-level context encoder creates a directed edge-labelled graph in which each utterance, enriched with context, is a node in the graph. The information from neighbor nodes is then aggregated and passed to a neural network in each node, resulting in an updated representation of each node that takes into account its neighbors. In DialogueGCN, node features are initialized with sequentially encoded feature vectors from the sequential context encoders, and edge weights are set using a similarity-based attention module. The utterance level emotion classification is turned into a problem of node classification in the graph. DialogueGCN achieves an F1-score of 64.18\% on the IEMOCAP dataset.

\subsection{Transformer}

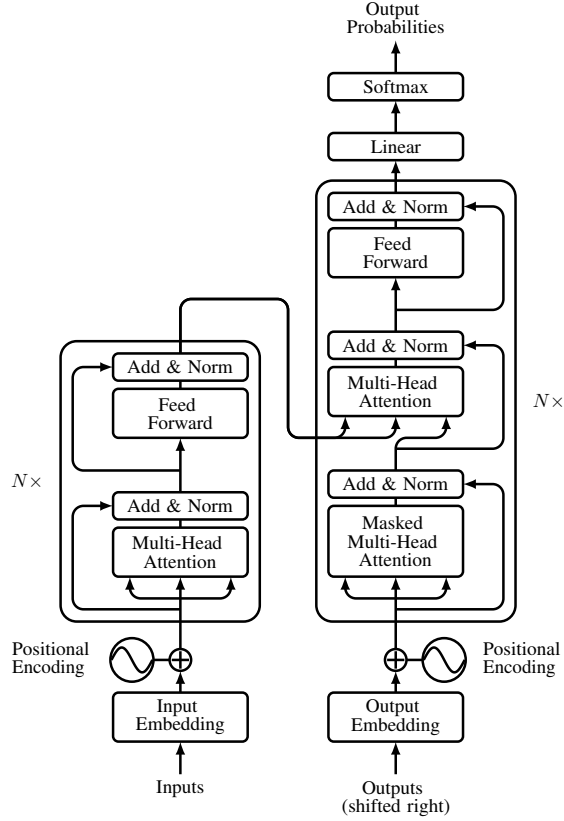
\begin{figure}[!ht]
	\begin{center}
		\resizebox{8cm}{11cm}{
			\begin{tikzpicture}
				\definecolor{emb_color}{RGB}{255,255,255}
				\definecolor{multi_head_attention_color}{RGB}{255,255,255}
				\definecolor{add_norm_color}{RGB}{255,255,255}
				\definecolor{ff_color}{RGB}{255,255,255}
				\definecolor{softmax_color}{RGB}{255,255,255}
				\definecolor{linear_color}{RGB}{255,255,255}
				\definecolor{gray_bbox_color}{RGB}{255,255,255}
				\draw[fill=gray_bbox_color, line width=0.046875cm, rounded corners=0.300000cm] (-0.975000, 6.455000) -- (2.725000, 6.455000) -- (2.725000, 1.305000) -- (-0.975000, 1.305000) -- cycle;
				\draw[fill=gray_bbox_color, line width=0.046875cm, rounded corners=0.300000cm] (3.775000, 9.405000) -- (7.475000, 9.405000) -- (7.475000, 1.305000) -- (3.775000, 1.305000) -- cycle;
				\draw[line width=0.046875cm, fill=emb_color, rounded corners=0.100000cm] (0.000000, 0.000000) -- (2.500000, 0.000000) -- (2.500000, -0.900000) -- (0.000000, -0.900000) -- cycle;
				\node[text width=2.500000cm, align=center] at (1.250000,-0.450000) {Input \vspace{-0.05cm} \linebreak Embedding};
				\draw[line width=0.046875cm, fill=emb_color, rounded corners=0.100000cm] (4.000000, 0.000000) -- (6.500000, 0.000000) -- (6.500000, -0.900000) -- (4.000000, -0.900000) -- cycle;
				\node[text width=2.500000cm, align=center] at (5.250000,-0.450000) {Output \vspace{-0.05cm} \linebreak Embedding};
				\draw[line width=0.046875cm, fill=add_norm_color, rounded corners=0.100000cm] (0.000000, 3.680000) -- (2.500000, 3.680000) -- (2.500000, 3.180000) -- (0.000000, 3.180000) -- cycle;
				\node[text width=2.500000cm, align=center] at (1.250000,3.430000) {Add \& Norm};
				\draw[line width=0.046875cm, fill=multi_head_attention_color, rounded corners=0.100000cm] (0.000000, 3.030000) -- (2.500000, 3.030000) -- (2.500000, 2.130000) -- (0.000000, 2.130000) -- cycle;
				\node[text width=2.500000cm, align=center] at (1.250000,2.580000) {Multi-Head \vspace{-0.05cm} \linebreak Attention};
				\draw[line width=0.046875cm] (1.250000, 3.030000) -- (1.250000, 3.180000);
				\draw[line width=0.046875cm, fill=add_norm_color, rounded corners=0.100000cm] (4.000000, 6.630000) -- (6.500000, 6.630000) -- (6.500000, 6.130000) -- (4.000000, 6.130000) -- cycle;
				\node[text width=2.500000cm, align=center] at (5.250000,6.380000) {Add \& Norm};
				\draw[line width=0.046875cm, fill=multi_head_attention_color, rounded corners=0.100000cm] (4.000000, 5.980000) -- (6.500000, 5.980000) -- (6.500000, 5.080000) -- (4.000000, 5.080000) -- cycle;
				\node[text width=2.500000cm, align=center] at (5.250000,5.530000) {Multi-Head \vspace{-0.05cm} \linebreak Attention};
				\draw[line width=0.046875cm] (5.250000, 5.980000) -- (5.250000, 6.130000);
				\draw[line width=0.046875cm, fill=add_norm_color, rounded corners=0.100000cm] (4.000000, 4.080000) -- (6.500000, 4.080000) -- (6.500000, 3.580000) -- (4.000000, 3.580000) -- cycle;
				\node[text width=2.500000cm, align=center] at (5.250000,3.830000) {Add \& Norm};
				\draw[line width=0.046875cm, fill=multi_head_attention_color, rounded corners=0.100000cm] (4.000000, 3.430000) -- (6.500000, 3.430000) -- (6.500000, 2.130000) -- (4.000000, 2.130000) -- cycle;
				\node[text width=2.500000cm, align=center] at (5.250000,2.780000) {Masked \vspace{-0.05cm} \linebreak Multi-Head \vspace{-0.05cm} \linebreak Attention};
				\draw[line width=0.046875cm] (5.250000, 3.430000) -- (5.250000, 3.580000);
				\draw[line width=0.046875cm, fill=add_norm_color, rounded corners=0.100000cm] (0.000000, 6.230000) -- (2.500000, 6.230000) -- (2.500000, 5.730000) -- (0.000000, 5.730000) -- cycle;
				\node[text width=2.500000cm, align=center] at (1.250000,5.980000) {Add \& Norm};
				\draw[line width=0.046875cm, fill=ff_color, rounded corners=0.100000cm] (0.000000, 5.580000) -- (2.500000, 5.580000) -- (2.500000, 4.680000) -- (0.000000, 4.680000) -- cycle;
				\node[text width=2.500000cm, align=center] at (1.250000,5.130000) {Feed \vspace{-0.05cm} \linebreak Forward};
				\draw[line width=0.046875cm] (1.250000, 5.580000) -- (1.250000, 5.730000);
				\draw[line width=0.046875cm, fill=add_norm_color, rounded corners=0.100000cm] (4.000000, 9.180000) -- (6.500000, 9.180000) -- (6.500000, 8.680000) -- (4.000000, 8.680000) -- cycle;
				\node[text width=2.500000cm, align=center] at (5.250000,8.930000) {Add \& Norm};
				\draw[line width=0.046875cm, fill=ff_color, rounded corners=0.100000cm] (4.000000, 8.530000) -- (6.500000, 8.530000) -- (6.500000, 7.630000) -- (4.000000, 7.630000) -- cycle;
				\node[text width=2.500000cm, align=center] at (5.250000,8.080000) {Feed \vspace{-0.05cm} \linebreak Forward};
				\draw[line width=0.046875cm] (5.250000, 8.530000) -- (5.250000, 8.680000);
				\draw[line width=0.046875cm, fill=linear_color, rounded corners=0.100000cm] (4.000000, 10.280000) -- (6.500000, 10.280000) -- (6.500000, 9.780000) -- (4.000000, 9.780000) -- cycle;
				\node[text width=2.500000cm, align=center] at (5.250000,10.030000) {Linear};
				\draw[line width=0.046875cm, fill=softmax_color, rounded corners=0.100000cm] (4.000000, 11.380000) -- (6.500000, 11.380000) -- (6.500000, 10.880000) -- (4.000000, 10.880000) -- cycle;
				\node[text width=2.500000cm, align=center] at (5.250000,11.130000) {Softmax};
				\draw[line width=0.046875cm] (1.250000, 0.600000) circle (0.200000);
				\draw[line width=0.046875cm] (1.410000, 0.600000) -- (1.090000, 0.600000);
				\draw[line width=0.046875cm] (1.250000, 0.760000) -- (1.250000, 0.440000);
				\draw[line width=0.046875cm] (5.250000, 0.600000) circle (0.200000);
				\draw[line width=0.046875cm] (5.410000, 0.600000) -- (5.090000, 0.600000);
				\draw[line width=0.046875cm] (5.250000, 0.760000) -- (5.250000, 0.440000);
				\draw[line width=0.046875cm] (0.350000, 0.600000) circle (0.400000);
				\draw[line width=0.046875cm] (-0.030000, 0.600000) -- (-0.014490, 0.629156) -- (0.001020, 0.657833) -- (0.016531, 0.685561) -- (0.032041, 0.711884) -- (0.047551, 0.736369) -- (0.063061, 0.758616) -- (0.078571, 0.778258) -- (0.094082, 0.794973) -- (0.109592, 0.808486) -- (0.125102, 0.818576) -- (0.140612, 0.825077) -- (0.156122, 0.827883) -- (0.171633, 0.826946) -- (0.187143, 0.822284) -- (0.202653, 0.813971) -- (0.218163, 0.802145) -- (0.233673, 0.786999) -- (0.249184, 0.768783) -- (0.264694, 0.747796) -- (0.280204, 0.724382) -- (0.295714, 0.698925) -- (0.311224, 0.671845) -- (0.326735, 0.643584) -- (0.342245, 0.614608) -- (0.357755, 0.585392) -- (0.373265, 0.556416) -- (0.388776, 0.528155) -- (0.404286, 0.501075) -- (0.419796, 0.475618) -- (0.435306, 0.452204) -- (0.450816, 0.431217) -- (0.466327, 0.413001) -- (0.481837, 0.397855) -- (0.497347, 0.386029) -- (0.512857, 0.377716) -- (0.528367, 0.373054) -- (0.543878, 0.372117) -- (0.559388, 0.374923) -- (0.574898, 0.381424) -- (0.590408, 0.391514) -- (0.605918, 0.405027) -- (0.621429, 0.421742) -- (0.636939, 0.441384) -- (0.652449, 0.463631) -- (0.667959, 0.488116) -- (0.683469, 0.514439) -- (0.698980, 0.542167) -- (0.714490, 0.570844) -- (0.730000, 0.600000);
				\draw[line width=0.046875cm] (6.150000, 0.600000) circle (0.400000);
				\draw[line width=0.046875cm] (5.770000, 0.600000) -- (5.785510, 0.629156) -- (5.801020, 0.657833) -- (5.816531, 0.685561) -- (5.832041, 0.711884) -- (5.847551, 0.736369) -- (5.863061, 0.758616) -- (5.878571, 0.778258) -- (5.894082, 0.794973) -- (5.909592, 0.808486) -- (5.925102, 0.818576) -- (5.940612, 0.825077) -- (5.956122, 0.827883) -- (5.971633, 0.826946) -- (5.987143, 0.822284) -- (6.002653, 0.813971) -- (6.018163, 0.802145) -- (6.033673, 0.786999) -- (6.049184, 0.768783) -- (6.064694, 0.747796) -- (6.080204, 0.724382) -- (6.095714, 0.698925) -- (6.111224, 0.671845) -- (6.126735, 0.643584) -- (6.142245, 0.614608) -- (6.157755, 0.585392) -- (6.173265, 0.556416) -- (6.188776, 0.528155) -- (6.204286, 0.501075) -- (6.219796, 0.475618) -- (6.235306, 0.452204) -- (6.250816, 0.431217) -- (6.266327, 0.413001) -- (6.281837, 0.397855) -- (6.297347, 0.386029) -- (6.312857, 0.377716) -- (6.328367, 0.373054) -- (6.343878, 0.372117) -- (6.359388, 0.374923) -- (6.374898, 0.381424) -- (6.390408, 0.391514) -- (6.405918, 0.405027) -- (6.421429, 0.421742) -- (6.436939, 0.441384) -- (6.452449, 0.463631) -- (6.467959, 0.488116) -- (6.483469, 0.514439) -- (6.498980, 0.542167) -- (6.514490, 0.570844) -- (6.530000, 0.600000);
				\draw[line width=0.046875cm, -latex] (1.250000, 3.680000) -- (1.250000, 4.680000);
				\draw[line width=0.046875cm, -latex] (5.250000, 6.630000) -- (5.250000, 7.630000);
				\draw[line width=0.046875cm, -latex] (5.250000, 9.180000) -- (5.250000, 9.780000);
				\draw[line width=0.046875cm, -latex] (5.250000, 10.280000) -- (5.250000, 10.880000);
				\draw[line width=0.046875cm, -latex] (1.250000, 0.000000) -- (1.250000, 0.400000);
				\draw[line width=0.046875cm, -latex] (1.250000, 0.800000) -- (1.250000, 2.130000);
				\draw[line width=0.046875cm, -latex] (5.250000, 0.800000) -- (5.250000, 2.130000);
				\draw[line width=0.046875cm, -latex] (5.250000, 0.000000) -- (5.250000, 0.400000);
				\draw[line width=0.046875cm] (0.750000, 0.600000) -- (1.050000, 0.600000);
				\draw[line width=0.046875cm] (5.450000, 0.600000) -- (5.750000, 0.600000);
				\draw[-latex, line width=0.046875cm, rounded corners=0.200000cm] (1.250000, 4.080000) -- (-0.750000, 4.080000) -- (-0.750000, 5.980000) -- (0.000000, 5.980000);
				\draw[-latex, line width=0.046875cm, rounded corners=0.200000cm] (1.250000, 1.530000) -- (-0.750000, 1.530000) -- (-0.750000, 3.430000) -- (0.000000, 3.430000);
				\draw[-latex, line width=0.046875cm, rounded corners=0.200000cm] (5.250000, 1.530000) -- (7.250000, 1.530000) -- (7.250000, 3.830000) -- (6.500000, 3.830000);
				\draw[-latex, line width=0.046875cm, rounded corners=0.200000cm] (5.250000, 4.480000) -- (7.250000, 4.480000) -- (7.250000, 6.380000) -- (6.500000, 6.380000);
				\draw[-latex, line width=0.046875cm, rounded corners=0.200000cm] (5.250000, 7.030000) -- (7.250000, 7.030000) -- (7.250000, 8.930000) -- (6.500000, 8.930000);
				\draw[-latex, line width=0.046875cm, rounded corners=0.200000cm] (1.250000, 1.730000) -- (0.312500, 1.730000) -- (0.312500, 2.130000);
				\draw[-latex, line width=0.046875cm, rounded corners=0.200000cm] (1.250000, 1.730000) -- (2.187500, 1.730000) -- (2.187500, 2.130000);
				\draw[-latex, line width=0.046875cm, rounded corners=0.200000cm] (5.250000, 1.730000) -- (4.312500, 1.730000) -- (4.312500, 2.130000);
				\draw[-latex, line width=0.046875cm, rounded corners=0.200000cm] (5.250000, 1.730000) -- (6.187500, 1.730000) -- (6.187500, 2.130000);
				\draw[-latex, line width=0.046875cm, rounded corners=0.200000cm] (1.250000, 6.230000) -- (1.250000, 7.230000) -- (3.250000, 7.230000) -- (3.250000, 4.680000) -- (4.312500, 4.680000) -- (4.312500, 5.080000);
				\draw[-latex, line width=0.046875cm, rounded corners=0.200000cm] (1.250000, 6.230000) -- (1.250000, 7.230000) -- (3.250000, 7.230000) -- (3.250000, 4.680000) -- (5.250000, 4.680000) -- (5.250000, 5.080000);
				\draw[-latex, line width=0.046875cm, rounded corners=0.200000cm] (5.250000, 4.080000) -- (5.250000, 4.680000) -- (6.187500, 4.680000) -- (6.187500, 5.080000);
				\draw[line width=0.046875cm, -latex] (1.250000, -1.500000) -- (1.250000, -0.900000);
				\draw[line width=0.046875cm, -latex] (5.250000, -1.500000) -- (5.250000, -0.900000);
				\draw[line width=0.046875cm, -latex] (5.250000, 11.380000) -- (5.250000, 11.980000);
				\node[text width=2.500000cm, anchor=north, align=center] at (1.250000,-1.500000) {Inputs};
				\node[text width=2.500000cm, anchor=north, align=center] at (5.250000,-1.500000) {Outputs \vspace{-0.05cm} \linebreak (shifted right)};
				\node[text width=2.500000cm, anchor=south, align=center] at (5.250000,11.980000) {Output \vspace{-0.05cm} \linebreak Probabilities};
				\node[anchor=east] at (-1.175000,3.880000) {$N\times$};
				\node[anchor=west] at (7.675000,5.355000) {$N\times$};
				\node[text width=2.000000cm, anchor=east] at (0.250000,0.600000) {Positional \vspace{-0.05cm} \linebreak Encoding};
				\node[text width=2.000000cm, anchor=west] at (6.750000,0.600000) {Positional \vspace{-0.05cm} \linebreak Encoding};
			\end{tikzpicture}
		}

	\end{center}
	\caption{The Transformer. The Encoder on the left and Decoder on the right display some similar components.
	}
	\label{transf}
\end{figure}

Despite the success of recurrent and graph neural networks in a wide variety of tasks including Emotion Recognition, a more powerful network model exists, constituting the current state-of-the-art: the Transformer \citep{vaswani2017attention}. The Transformer is also better at capturing long-term dependencies than gated RNNs, due to its shorter path of information flow, which is useful when modelling several utterances in Emotion Recognition in Conversations.  It does not resort to recurrence or convolutions. It is also more parallelizable and needs less training time. It is composed of an Encoder and Decoder, which can be visualized in Figure \ref{transf}.

Each Encoder layer is composed of two sub-layers. The first is a so called self-attention layer for building the context of the sequence that calculates the relevance of the tokens in the sequence for each given token. 

The self-attention mechanism takes as input three matrices, the Query, Key, and Value matrices. These are computed from the dot product between the input sequence matrix and weight matrices learned in the training phase. It then computes a vector of attention scores, as described in the following equation:

\begin{equation}\label{eq8}
	Attention(Q, K, V )=softmax\left(\frac{QK^T}{\sqrt{k}}\right)V,
\end{equation}

in which $Q \in \mathbb{R}^{N\times m}$, $K \in \mathbb{R}^{N\times m}$ and $V \in \mathbb{R}^{N\times m}$ are the Query, Key, and Value matrices, respectively, and $m$ is the model output dimension. The dot product of the Query matrix with the Key matrix represents scores that are then multiplied by the Value matrix to keep intact the values of relevant words and drown-out the irrelevant ones. The $\frac{1}{\sqrt{k}}$ helps stabilizing the gradients.

The self-attention layer uses a multi-head attention mechanism so that the attention score vectors do not exclusively give a high probability score for its corresponding token, defined by the following equations:

\begin{equation}\label{eq9}
	MultiHead(Q, K, V ) = Concat(head_1, ..., head_h)W^{O}, 
\end{equation}
where
\begin{equation}\label{eq10}
	head_i = Attention(QW_i^Q , K W_i^K , V W_i^V ),
\end{equation}

in which $W_i^Q  \in \mathbb{R}^{m\times k}$ , $W_i^K  \in \mathbb{R}^{m\times k}$ and $W_i^V  \in \mathbb{R}^{m\times v}$ are projection matrices and $W^O  \in \mathbb{R}^{hv\times m}$, being $h$ the number of heads and $k$=$v$=$m/h$.

The output of the self-attention layer is fed to the second layer, a fully connected feed-forward network, with two linear transformations with a ReLU activation in between, applied separately and in the same way to each position, allowing for parallelization. 

A residual connection is applied around each of the two sub-layers, followed by layer normalization.

Each Decoder layer is composed of three sublayers. The first is a masked self-attention layer similar to the self-attention layer described mathematically in Equations \ref{eq9} and \ref{eq10}, but in which each position in the decoder attends to all positions in the decoder up to and including that position, which is attained by masking out values in the softmax. The second is an “encoder-decoder attention” layer, described mathematically in the same equations, where K and V come from the output of the encoder and Q comes from the output of the decoder’s masked self-attention layer, mimicking the typical encoder-decoder attention mechanisms in Seq2Seq models. The last layer is a feed-forward neural network, similar to the Encoder layer.

Similar to the Encoder, a residual connection is applied around each of the two sub-layers, followed by layer normalization.

Positional encodings are added to the embeddings that are fed to the Encoder and Decoder to introduce information about the position of the tokens in the sequence.

The most common use of the Transformer is as the backbone for a fine-tuned pre-trained transformer-based language model, such as BERT \citep{devlin2018bert}, described in subsection \ref{embeddings}. These language models are trained with large amounts of data to perform many tasks and only need to be fine-tuned for the specific task at hand. In the case of a classification task such as Emotion Recognition, one just needs to add a linear layer or a more complex neural network architecture on top of the language model. The training is performed with supervised learning, resorting to backpropagation algorithms.

\subsubsection[Knowledge-Enriched Transformer]{Knowledge-Enriched Transformer \citep{zhong2019knowledge}}

The invention of the Transformer \citep{vaswani2017attention} led to new state-of-the-art results in several Natural Language Processing tasks. The Knowledge-Enriched Transformer (KET) uses self-attention to model context and response using the Transformer, that a has shorter path of information flow than gated RNNs, overcoming the difficulty in capturing long-term dependencies. Conversations are modeled in the entire Transformer as a single input. 

First, concepts are retrieved for each word in the conversation using an external knowledge base, a graph of concepts, then the concept representation is computed using a dynamic context-aware graph attention mechanism over the concepts that are then combined with the input sentence embeddings. Self-attention is used to learn utterance representations individually and hierarchical self-attention is applied to the context to learn the context representation.  Finally, the encoder and decoder attention mechanism is applied, followed by max pooling and a linear layer. KET achieves an F1-score of 59.56\% on the IEMOCAP dataset.

\subsubsection[CESTa]{CESTa \citep{wang2020contextualized}} 

The CESTa or Contextualized Emotion Sequence Tagging approaches ERC as a task of sequence tagging, choosing the set of tags with the highest likelihood for the whole utterance sequence at once, by using a Conditional Random Field (CRF) \citep{lafferty2001conditional}. In order to capture long-range global context, utterance representations are generated by a multi-layer Transformer encoder. These are then fed to a bi-LSTM encoder that captures self and inter-speaker dependencies, resulting in contextualized representations of utterances that are then passed to the CRF layer. It achieves an F1-score of 67.10\% in the IEMOCAP dataset. 

\subsubsection[DialogXL]{DialogXL \citep{shen2020dialogxl}}

DialogXL is an adaptation of the pre-trained language model XLNet \citep{yang2019xlnet}, which is based on the Transformer-XL \citep{dai2019transformer}, for Emotion Recognition in Conversations. 
It replaces XLNet’s segment recurrence with a memory utterance recurrence to leverage historical utterances. These utterances' hidden states are stored in a memory bank to reuse them while identifying a query utterance. The approach also replaces XLNet's vanilla self-attention proposing a dialog-aware self-attention to grasp useful intra- and inter-speaker dependencies. This is composed of four types of self-attention: global and local self-attention for different sizes of receptive fields, and speaker and listener self-attention for intra- and inter-speaker dependencies. DialogXL achieves an F1-score of 65.94\% on the IEMOCAP dataset.

\subsection{Embeddings from Pre-Trained Transformer Language Models}
\label{embeddings}

Text features must represent words in a way that is useful for the models, in this case, numeric vectors referred to as word embeddings. These embeddings yield similar representations for similar words. 

Static embeddings \citep{mikolov2013efficient, pennington2014glove, mikolov2017advances} are obtained based on the co-occurrence of adjacent words and have a fixed representation for the words, not taking into account their context.

Contrarily, contextual embeddings include information from the context in the word representation, a representation that differs according to the occurrence of the word. The only disadvantage of this kind of representation is that we cannot use pre-trained embeddings as in static embeddings so there is an inference time for generating the embeddings. The models are pre-trained with large amounts of data, in order to enable generating good embedings.
A first approach, Embeddings from Language Models (ELMo) \citep{peters2018deep} is based on a stack of two Bi-LSTMs, leveraging the full context of a word. It provides a context-free representation of the word along with context information of the sense of the word and its syntax.
The second approach, Bidirectional Encoder Representations from Transformers (BERT) \citep{devlin2018bert}, built from a Transformer encoder stack, leads to state-of-the-art results in multiple Natural Language Processing tasks. The input given to BERT is the sum of three embeddings, the token embedding corresponding to the sequence of words, the segment embedding containing information on which sentence each token belongs to, and the position embedding, defining the position of each token in the sequence and distance to other tokens taking into account the order of the sequence. An improvement upon BERT, RoBERTa \citep{liu2019roberta} is trained with more data and for a longer period of time.
Given that context is key for text Emotion Recognition in Conversations, contextual word embeddings are more adequate for the task than static embeddings, as can be seen in the higher performance of works that resort to these embeddings, namely resultant from Pre-trained language models, such as BERT-based embeddings. The latter are also more tailored to deal with common sense, informal language, and sarcasm, ERC challenges described in Section \ref{challenges}, since they are pre-trained on large amounts of text.

\subsubsection [COSMIC]{COSMIC \citep{ghosal2020cosmic}}

Besides choosing appropriate utterance representations and classifier architectures, there are other factors that can be leveraged for the task of ERC. One such factor is the commonsense knowledge of the interlocutors that plays a central role in inferring the latent variables of a conversation, such as speaker state and intent.

COSMIC extracts commonsense features resorting to COMET \citep{bosselut2019comet}, a commonsense transformer encoder-decoder model trained on the task of generative commonsense knowledge construction that uses GPT \citep{radford2019language} as its generative model.
The approach uses RoBERTa for context-independent feature extraction, passes each utterance through COMET's encoder, and extracts the activations from the final time step. 
COSMIC maintains a context state and attention vector which are always shared between the participants of the conversation.
Five GRUs are used to model context state, internal state, external state, intent state, and emotion state, being the latter used for emotion classification.
This work achieves an F1-score of 65.28\% in the IEMOCAP dataset.

\subsubsection[Psychological]{Psychological \citep{li2021past}} 

Psychological also leverages COMET as the commonsense knowledge base and resorts to RoBERTa as the utterance-level encoder. It proposes SKAIG as a conversation-level encoder, a locally connected graph, where the targeted utterance receives information from that past and future context and is also self-connected. 

Assuming that the influence of an utterance on contextual utterances is locally effective, the targeted node is connected with contextual nodes in a window of a given size of utterances in the past and future.
Knowledge from COMET is introduced to enrich the edges with different relations. To propagate information through SKAIG a Graph Transformer \citep{shi2020masked} is used.
Finally, a linear unit predicts the emotion distributions. 
This work achieves an F1-score of 66.96\% in the IEMOCAP dataset.

\subsubsection[DAG-ERC]{DAG-ERC \citep{shen2021directed}}

Graph-based methods tend to neglect distant utterances and sequential information. On the other hand, recurrence-based methods leverage those but tend to update the query utterance’s state with limited information from the nearest utterances. Combining both graph and recurrence structures explores their advantages and mitigates their drawbacks. This is the idea behind DAG-ERC, which regards each conversation as a directed acyclic graph (DAG), a combination of both structures. DAG-ERC is based on the DAGNN \citep{thost2021directed} architecture, with two improvements: a relation-aware feature transformation that gathers information based on speaker identity and a contextual information unit that enhances the information of historical context. RoBERTa-Large is used as the feature extractor. 
It achieves an F1-score of 68.03\% on the IEMOCAP dataset, outperforming the previous approaches.

\subsection{ERC within a Generative Framework}

With the advent of Generative Large Language Models (LLMs), such as the GPT series \citep{radford2018improving, radford2019language, brown2020language} and Llama \citep{touvron2023llama}, classification tasks started to be reformulated within generative frameworks. The idea is to prompt the LLM with the utterance to be classified and other relevant additional input and request an emotion label for the utterance. Open-source LLMs can also be fine-tuned and even pre-trained. 

\subsubsection[InstructERC]{InstructERC \citep{lei2023instructerc}}

In InstructERC the prompt consists of the utterance to be classified, the ERC instruction, the conversational context, the set of possible classification labels, and similar utterances together with its emotion labels. The LLM is pre-trained with a speaker identification task and fine-tuned both with the main ERC task and an emotion influence prediction task to improve overall performance. Experiments were made with several LLMs, and the best results were obtained with LLama2. InstructERC achieves 71.39\% in the IEMOCAP dataset, outperforming the previous approaches.

\section{Advisable ERC Practices}
\label{practices}

This section provides useful methods for better ERC frameworks, from different ways of dealing with subjectivity in emotion annotation and modelling, to several ways of dealing with the typically unbalanced ERC datasets.

\subsection{Learning Subjectivity}
\label{annotation}

All human annotation tasks involve some degree of uncertainty, that can stem amongst other factors from the subjectivity inherent to the annotator. Since Emotion Recognition is amongst the most subjective tasks \citep{uma2021learning} and that subjectivity is the main source of uncertainty in Emotion Recognition annotation, we address uncertainty as subjectivity in this survey \citep{rizos2020average}.  A good annotation practice is to resort to several annotators to avoid biasing the results towards a single annotator's subjectivity. This will naturally yield different labels from each annotator, the so called annotation disagreement. The more appropriately one deals with the disagreement in the annotated labels the more reliable will the data and classifiers be. Several techniques for dealing with disagreement are described in this section. They are divided into techniques that address subjectivity by adapting the labels, towards an estimate of a gold label, and techniques that embrace subjectivity by adapting the classifier architectures to deal with the subjectivity of the labels, leveraging such property.

\subsubsection{Addressing Subjectivity}
\label{adressing}

For categorical labels, considering that a gold or true label exists for each sample, the simple way to obtain it would be through majority voting, and choosing the label with the most annotations. However, this does not account for the different annotator skill levels or sample inherent difficulty. 

Thus, a common process is to calculate an inter-annotator agreement score to find low agreement data. A very simplistic way of calculating inter-annotator agreement for categorical labels would be to use the percent agreement, dividing the number of agreement classifications by the total number of classifications. However, non-expert annotators do not always know which label to use so they sometimes just guess, resulting in a random label. Cohen pointed out that there is some level of agreement in these random labels and developed Cohen's Kappa to take that factor into account \citep{mchugh2012interrater}. The related Fleiss' Kappa \citep{fleiss1971measuring} is the adaptation of Cohen's Kappa for 3 or more annotators.

The low agreement data can then be discarded or trained on and evaluated separately from the high agreement data. It can be considered random noise if it does not reveal the existence of annotator bias \citep{uma2021learning}. 

For interval labels, weighted fusion of data can be performed with approaches such as the Evaluator Weighted Estimator (EWE) \citep{grimm2005evaluation}, which weights each rater's annotations based on its rater-specific inter-reliability scores, or the Weighted Trustability Evaluator (WTE) \citep{hantke2016introducing}, which is instead based on each rater's performance consistency.

\subsubsection[Embracing Subjectivity]{Embracing Subjectivity \citep{rizos2020average}}

The annotator (dis)agreement can be used as privileged information for master-class learning, being viewed as samples' additional information \citep{eyben2012multitask} to facilitate learning in the training phase, that might or might not be required for making predictions during testing, for example, to weigh positively the loss of the corresponding samples. In this setting, it is useful to learn the meaning of high-rater disagreement for the particular dataset in use.

Instead of considering the existence of a hard label, as in subsubsection \ref{adressing}, one can calculate the distribution of label annotations to define a soft label distribution per sample, which encodes the subjectivity for each sample and allows for the classifier to learn label correlations.
Another approach is to model the annotations of each particular rater, through an ensemble of models. It allows to give more importance to expert raters rather than novices or spammers.
The addition of an "unsure" label would facilitate the annotator's workload and also make the model learn the ambiguity ground-truth, although not having to predict "unsure" labels during testing.
Furthermore, methods to understand whether a sample is mislabelled by certain raters or inherently ambiguous could inform the model and annotators during active learning processes.

In ERC, the context of the interactions is not always explicit and the dialogues contain a variety of topics that can exacerbate subjectivity even more, making it much more advisable to embrace subjectivity in this domain.

\subsection{Dealing with Unbalanced Data}
\label{unbalanced}

All the benchmark datasets used for ERC are unbalanced, some of them are severely unbalanced: the unbalance ratio varies from 1:3 up to 1:1156. In the presence of these unbalances, most classifiers tend to favor the majority classes which, in these datasets, corresponds to the neutral emotion class. Hence, unless the unbalance issue is addressed, most models will be trained and evaluated based on the neutral class results, which is far from intended. All classes should be equally relevant in ERC, and even when Accuracy values look acceptable, a more detailed analysis usually reveals a very poor performance in minority classes.

\subsubsection{Metrics for Unbalanced Data}

The performance metrics used for Emotion Recognition in Conversations are Accuracy, Precision, Recall, and F-score or F1, the harmonic mean of Precision and Recall. These are commonly used metrics in many classification applications. One can consider the weighted version of these metrics, that take into account the relative frequencies of each class, or micro-averaging, in which all samples equally contribute to the final averaged metric. Since both weighted versions and micro-averaging depend more on the classes with a bigger number of items, those are not good indicators of the performance of the minority classes. Therefore, those should not be used when the classes have the same importance and unequal frequencies. In unweighted macro-averaging, the metric is computed for each class, and the results are averaged over all the classes. By maximizing an unweighted metric, one is maximizing the performance of the model in correctly classifying all classes regardless of the number of items they have. 

ERC is clearly a task that deals with real-world unbalanced data. As such, a proper evaluation should always be performed on a test set that maintains the real-world unbalance of the data and using unweighted metrics such as macro F-score. Ideally, performance results on each of the individual classes should also be presented. Accuracy, micro-averaging, and weighted metrics are mostly irrelevant in what concerns the performance of the models since they are mostly dependent on the majority class, usually neutral. Unweighted metrics that are not affected by dataset unbalance, such as Recall, should also be avoided, unless used in conjunction with those that are, such as Precision.  

\subsubsection{Balancing Techniques}

Besides the use of appropriate metrics, another way of dealing with dataset unbalance is to apply balancing techniques to the training set. Common such techniques are random under-sampling, removing observations from the majority class, and random over-sampling, adding copies to the minority class. More elaborate techniques, such as Smote - Synthetic Minority Oversampling Technique \citep{chawla2002smote}, usually improve the results.

Balancing a training set has its advantages, but it is of utmost importance that the test set maintains the original set balance. Otherwise, metrics such as Precision and F-score are artificially improved and the model is unlikely to be effective in an unbalanced real-world dataset. It should be noted that Recall is not affected when training and testing on an artificially balanced dataset so ignoring Precision and looking only at Recall as an evaluation metric will not be a good indicator for real-world performance of a model when there is a real-world imbalance among classes.

\subsubsection{Few-Shot Learning}

Few-Shot Learning \citep{wang2020generalizing} is also an efficient method for dealing with unbalanced data. Contrary to supervised learning, it does not require a high number of training examples. Instead of learning to generalize class identities from a training set to a test set, the Few-Shot Learning models learn to discriminate the similarities and differences between classes, by training a function that predicts similarity.

\section{Systematic Review of ERC Works}
\label{tables}
In Table \ref{ercmul}, a plethora of works on ERC, since 2017, are displayed along with the methods they use and their reported performance.

\begin{longtable}{|llcccc|}
	\caption{Summary of publications on Emotion Recognition in Conversations, reporting the methods used along with their performance across several datasets. A * indicates that the work is multimodal.}\\
	\hline
	\textbf{Author, Year}  & \textbf{Methods}     & \textbf{IEMO} & \textbf{MELD} & \textbf{ENLP} &  \textbf{DD} \\
	
	\hline
	
	&\textbf{GRU}&\textbf{w-F1(6)}&\textbf{w-F1}&\textbf{w-F1}&\textbf{m-F1}\\
	\hline
	Jiao et al \cite{jiao2020real}*& Bi-GRU, Attention &62.7 &58.1 &&\\

	Majumder et al \cite{majumder2019dialoguernn}*& Bi-GRU, Attention &62.75 &&&\\

	Lu et al \cite{lu2020iterative}&Bi-GRU, Attention&64.37 &60.72 &&\\

	\hline
	&\textbf{LSTM}&&&&\\
	\hline
	Partaourides et al \cite{partaourides2020self}&Bi-LSTM, GRU, Attention &53.0 &&&\\
	Poria et al \cite{poria2017context}* & LSTM &54.95&&&\\

	Xing et al \cite{9128015}*&Bi-LSTM, GRU, Attention &64.30&60.45&&\\
	\hline
	\textbf{Author, Year}  & \textbf{Methods}     & \textbf{IEMO} & \textbf{MELD} & \textbf{ENLP} &  \textbf{DD} \\
	
	\hline
	Hu et al \cite{hu2021dialoguecrn}&Bi-LSTM, Attention&66.20&58.39&&\\
	\hline

	&\textbf{Memory Network}&&&&\\
	\hline
	Hazarika et al \cite{hazarika2018conversational}* & Memory Network, GRU, Attention&56.13&&&\\
	Hazarika et al \cite{hazarika2018icon}*& Memory Network, GRU, Attention &58.54 &&&\\
	Lai et al \cite{lai2020different}*&Memory Network, GRU, Attention&62.43&&&\\
	
	\hline
	&\textbf{Graph Neural Network (GNN)}&&&&\\
	\hline
	
	Ghosal et al \cite{ghosal2019dialoguegcn}&Graph Neural Network, GRU&64.18&58.10&&\\
	
	Ishiwatari et al \cite{ishiwatari2020relation}&Graph Neural Network, Attention&65.22&60.91&34.42&54.31\\
	
	Hu et al \cite{hu2021mmgcn}*&Graph Neural Network&66.22&58.65&&\\

	Sheng et al \cite{sheng2020summarize}&Graph Neural Network, Bi-LSTM, Attention &66.61&58.45&&\\

	Hu et al \cite{hu2022mm}* & Graph Neural Network, Bi-GRU&68.18&59.46&&\\

	\hline
	Li et al \cite{li2021quantum}*&\textbf{Quantum-inspired Neural Network }&59.88&58.00&&\\
	\hline
	&\textbf{Transformer}&&&&\\
	\hline
	
	Zhong et al \cite{zhong2019knowledge}& Transformer, Knowledge Base (KB) &59.56&58.18&34.39&53.37\\

	Zhang et al \cite{zhang2020knowledge}&Transformer&61.43&58.97&35.59&54.71\\
	Zhu et al \cite{zhu2021topic} & Transformer, KB &62.81&68.23&43.12&58.47   \\
	Khang and Cho \cite{10287935} & Transformer, GNN, Few-Shot Learning &63.16&&&52.83\\
	Sun et al \cite{sun2021discourse} &Transformer, Bi-LSTM, GNN & 64.10&64.22&36.38&\\

	Li et al \cite{li2020hitrans}&Transformer&64.50&61.94&36.75&\\
	
	Ghosal et al \cite{ghosal2020cosmic}&Transformer, GRU, KB &65.28&65.21&38.15&58.48\\
	
	Tu et al \cite{9956021}&Transformer, GRU, KB &65.69&66.12&39.47&\\
	Shen et al \cite{shen2020dialogxl}&Transformer&65.94&62.41&34.73&54.93\\
	
	Li et al \cite{li2022contrast} &Transformer, Contrastive Learning \citep{gao-etal-2021-simcse}&66.18&69.70&49.08&56.46\\
	Zhang et al \cite{10096161} &Transformer, Attention&66.35&&38.93&61.22\\    
	Li et al \cite{li2021past} &Transformer, GNN, KB&66.96&65.18&38.88&59.75\\

	Xie et al \cite{xie2021knowledge} &Transformer, KB, Attention&66.98&63.24&&57.30\\
	
	Wang et al \cite{wang2020contextualized} &Transformer, Bi-LSTM, CRF &67.10&58.36&&63.12\\

	Tu et al \cite{tu-etal-2023-context}&Transformer, Constrastive Learning&67.16&66.21&40.23&60.96\\
	
	Zhang et al \cite{zhang-etal-2023-dualgats}&Transformer, GNN&67.68&66.90&40.69&61.84\\
	Shen et al \cite{shen2021directed}&Transformer, DAGNN, GRU&68.03&63.65&39.02&59.33\\
	Yang et al \cite{10418539}&Transformer, GNN, LSTM, KB&68.31&66.25&40.23&60.21\\
	
	Tu et al \cite{tu-etal-2023-empirical}&Transformer, GNN, Contrastive Learning&68.49&65.34&39.20&\\
	Quan et al \cite{10120992}&Transformer, GNN&68.50&63.80&39.19&59.39\\
	Za et al \cite{10447592}&Transformer, KB, GNN, GRU&68.53&63.92&39.56&59.78\\
	
	Yang et al \cite{yang2021hybrid}&Transformer, GNN&68.73&66.18&46.11&59.76\\
	Su et al \cite{10541059}&Transformer - Generative LLM, GNN, LSTM&68.90&67.50&&\\
	Liu et al \cite{liu-etal-2022-dialogueein}&Transformer, Attention &68.93&65.37&&\\
	
	Liang et al \cite{liang-etal-2022-page}&Transformer, GNN&68.93&64.17&40.05&64.18\\
	
	Wu et al \cite{10356052}*&Transformer, Bi-LSTM, Attention&69.09&&&\\
	Duong et al \cite{10378668}&Transformer, GNN&69.10&63.82&39.85&\\
	Mao et al \cite{mao2020dialoguetrm}*& Transformer &69.23&63.55&&\\

	Zhang and Li \cite{zhang-li-2023-cross}*&Transformer, GNN&69.6&62.3&&\\

	Song et al \cite{song-etal-2022-supervised}&Transformer, Contrastive Learning&69.74&67.25&40.94&\\
	
	Yang et al \cite{10040720}&Transformer, Adapter, Contrastive Learning&69.81&65.70&38.75&62.51\\
	Xu and Yang \cite{10446410}&Transformer, Attention, Contrastive Learning &69.91&67.58&40.79&60.70\\
	Zhang et al \cite{zhang2023dialoguellm}*&Transformer - Generative LLM&69.93&71.90&40.05&\\  
	Li et al \cite{10081075}*&Transformer, GNN&70.00&58.94&&\\
	Hou et al \cite{hou-etal-2023-enhancing}&Transformer, GNN, Contrastive Learning&70.16&66.65&41.06&62.19\\
	Yang et al \cite{10135132}&Transformer, Variational Autoencoder \citep{kingma2013auto}&70.22&65.94&&62.14\\
	Tu et al \cite{tu-etal-2023-training}&Transformer, GNN&70.43&66.19&40.51&\\
	Yu et al \cite{yu-etal-2024-emotion}&Transformer, Contrastive Learning&70.41&67.12&40.24&\\
	
	Yun et al \cite{yun-etal-2024-telme}*&Transformer&70.48&67.37&&\\
	
	Li et al \cite{li-etal-2023-joyful}*&Transformer, GNN&71.03&61.77&&\\
	Li et al \cite{10502283}*&Transformer&71.04&66.70&&\\
	Yao and Shi \cite{10447720}*&Transformer, GNN, LSTM, Bi-GRU&71.21&66.25&&\\
	Lei et al \cite{lei2023instructerc}&Transformer - Generative LLM&71.39&69.15&41.37&\\
	Zhang et al \cite{10417143}*&Transformer, GNN&71.60&62.30&&\\
	\hline
	\textbf{Author, Year}  & \textbf{Methods}     & \textbf{IEMO} & \textbf{MELD} & \textbf{ENLP} &  \textbf{DD} \\
	
	\hline
	Wei et al \cite{10094596}*&Transformer, GNN&71.60&63.92&&\\
	Li et al \cite{li2022emocaps}* &Transformer, Bi-LSTM&71.77&64.00&&\\
	Hu et al \cite{hu2024unimeec}*&Transformer - Generative LLM, GNN&74.83&68.75&&\\
	Hong et al \cite{detectivenn}*&Transformer, Bi-GRU&76.01&&40.78&\\
	Jiang et al \cite{9999285} &Transformer, KB&&58.66&35.77&54.82\\
	Lee and Choi \cite{lee2021graph} &Transformer, GNN&&65.36&39.24&61.91\\
    	Chen et al \cite{10094810}&Transformer, GRU&&66.08&39.48&\\
	Zhao et al \cite{zhao-etal-2022-mucdn} &Transformer, GRU, Attention &&65.37&40.09&\\

	Sun et al \cite{sun2021discourse} &Transformer, GNN, Bi-LSTM  &&64.22&36.38& \\
	
	Jian et al \cite{10446226} &Transformer, GNN&&67.60&&\\

	\hline
	\label{ercmul} 

\end{longtable}

Several authors resorted to gated neural networks, which can capture dependencies between words and utterances. This effort led to high reported results on the benchmark datasets. Some authors presented works using graph neural networks and the performances obtained are comparable to the ones obtained with gated neural networks. The best performances, however, were yielded when using transformers or transformer-based models, that better capture dependencies in long utterances, combined with the aforementioned state-of-the-art deep-gated and graph-based network architectures for context modelling.  This may reflect the promising role that transformers and pre-trained transformer-based language models play in Natural Language Processing and Emotion Recognition in Conversations and the suitability of recurrent gated and graph-based neural network architectures, as elaborated in previous sections. 

\subsection{ERC Challenges Revisited}
\label{revisited}
Practically all works leveraged the information of preceding and sometimes also subsequent \citep{li2021past} utterances, resorting to gated or graph neural networks for context modelling of the utterances that were represented by embeddings from fine-tuned pre-trained language models. Another viable way to perform context modelling is to feed several appended utterances to the pre-trained language model \citep{pereira-etal-2023-context}. For speaker-specific modelling, also gated \citep{majumder2019dialoguernn}, graph \citep{ghosal2019dialoguegcn} neural networks and a combination of both \citep{shen2021directed} architectures were used. Emotion dynamics modelling was less explored. One work considered a GRU to model the emotion dynamics of each party \citep{majumder2019dialoguernn}.
Some works considered a knowledge base \citep{zhong2019knowledge, ghosal2020cosmic, li2021past, 10447592} to aid in interpreting the meaning behind commonsense knowledge. The majority of the works resorted to fine-tuning large pre-trained language models to obtain embeddings, which also aids in capturing the meaning behind commonsense knowledge expressions and informal language. 

\section{Directions for Future Research}
\label{future} 

We now present suggestions for future work directions. 

The first set of directions concerns potentiating the applicability of ERC modules for real-life scenarios. It comprises real-time ERC, recognizing emotion causes, and multilingual ERC, all elaborated in Section \ref{challenges}. There are plenty of research opportunities regarding annotation and modelling efforts. 

Further exploring mixed emotions with multi-label datasets and classifiers is also promising. Moreover, training the same model with different datasets by matching emotions from one dataset to another improves the generalization capabilities of the classifiers.

Some surveyed works resorted to knowledge bases. These are, however, not specific to emotions, which would be a useful extension.

Most datasets are unbalanced, motivating the need to leverage techniques and use appropriate performance metrics to address this unbalance, as elaborated in subsection \ref{unbalanced}.

Concerning subjectivity uncertainty in annotations, we highlight using inter-annotator (dis)agreement measures to weight or filter the annotator's opinions, incorporating these measures as additional information to the classifiers, and considering soft label distributions per sample.

Finally, we encourage further tackling interpretability, as elaborated in Section \ref{challenges}, since most works focus on performance.

\section{Conclusion}

\label{conclusion}

Research in Emotion Recognition in Conversations is advancing at a high pace and novel application scenarios pose new challenges and opportunities. 
Although research in Emotion Recognition in Conversations has come a long way, there is still much room for improvement, as it can be seen by the performance of the classifiers on the benchmark datasets and the several unexplored directions put forward in this section. While we described how current work has addressed several challenges, we also pointed out the opportunities that partly unaddressed challenges constitute. 

As main contributions of this survey, we presented partly addressed challenges for ERC, such as recognizing emotion causes, dealing with different taxonomies across datasets, multilingual ERC, and interpretability with associated future work directions. We compiled an extensive list of Deep Learning works in ERC, being the first to simultaneously report their methods, modalities, and performance across various datasets.
Finally, our descriptions of Deep Learning methods in the context of ERC provided insights into the suitability of each method for this task, which were not given in previous surveys. 

The survey highlights the advantage of leveraging techniques to address unbalanced data, the exploration of mixed emotions, and the benefits of incorporating annotation subjectivity in the learning phase.

Our survey relies on established benchmark datasets. While this aids in comparing different works, it also constitutes a limitation since benchmarking on these datasets may not generalize well to domain-specific applications such as different industries or languages. We also just briefly mention some challenges of real-world applications, not elaborating on the scalability or performance of models in such settings. So our findings may be highly relevant to academic benchmarks but less relevant to domain-specific applications.

Finally, important ethical aspects pertaining to Emotion Recognition are now presented. These aspects are, for example, and not limited to, whether an Emotion Recognition module should be developed or used for a certain purpose, which data to collect and the subjects behind the data, diversity and inclusiveness, privacy and control, and possible biases and misuses of the application \citep{mohammad2022ethics}. Research in these directions will benefit the community with better Emotion Recognition in Conversation modules for current and novel applications.

\section{Acknowledgments}

This work was supported by Fundação para a Ciência e a Tecnologia (FCT), through Portuguese national funds, Ref. UIDB/50021/2020, DOI: 10.54499/UIDB/50021/2020 and Ref. UI/BD/154561/2022 and the Portuguese Recovery and Resilience Plan through project C645008882-00000055 (Responsible.AI). Deep Learning TikZ images were based on contributions from Mark Wibrow, user121799, J. Leon V on StackExchange, and Renato Negrinho on GitHub.

\end{document}